\pdfoutput=1

\documentclass[11pt]{article}

\usepackage[final]{acl}

\usepackage[skins,most]{tcolorbox} 
\tcbuselibrary{breakable} 
\newtcolorbox[auto counter, number within=section, list type=subsubsection, list inside=toc]{sectionbox}[2][]{
colback=white!98!gray, colframe=black, 
colbacktitle=white!90!gray, coltitle=black, 
fonttitle=\bfseries,
title={#2}, 
list entry={Comment \thetcbcounter\quad}
}

\usepackage{booktabs}
\usepackage{subcaption}
\usepackage{times}
\usepackage{latexsym}
\usepackage{amsmath}
\usepackage{algorithm}
\usepackage{algpseudocode}
\usepackage{xcolor}
\usepackage{color}
\usepackage{listings}
\definecolor{customTeal}{RGB}{0, 128, 128}
\definecolor{emphasisColor}{RGB}{255, 0, 0} 
\usepackage{tabularx}
\usepackage{ragged2e}
\newcolumntype{Y}{>{\RaggedRight\arraybackslash}X} 
\usepackage[skins,most]{tcolorbox}
\usepackage{colortbl}
\usepackage{multirow} 
\usepackage{booktabs}
\usepackage{wrapfig}

\usepackage{pifont}  
\definecolor{customgreen}{HTML}{16C47F}  
\definecolor{customred}{HTML}{C62300}   

\newcommand{\cmark}{\textcolor{customgreen}{\ding{51}}}  
\newcommand{\xmark}{\textcolor{customred}{\ding{55}}}   

\usepackage{float}

\usepackage{caption}

\usepackage[T1]{fontenc}

\usepackage[utf8]{inputenc}

\usepackage{microtype}

\usepackage{inconsolata}

\usepackage{enumitem} 
\usepackage{lipsum}
\usepackage{graphicx}

\NewDocumentCommand{\yi}
{ mO{} }{\textcolor{blue}{\textsuperscript{\textit{May}}\textsf{\textbf{\small[#1]}}}}

\title{\raisebox{-0.15cm}{\includegraphics[width=0.8cm]{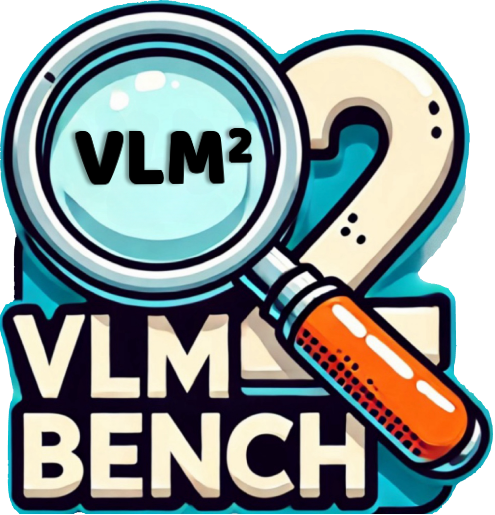}}VLM2-Bench: A Closer Look at How Well VLMs Implicitly Link \\ Explicit Matching Visual Cues}

\DeclareSymbolFont{extraup}{U}{zavm}{m}{n}
\DeclareMathSymbol{\vardiamond}{\mathalpha}{extraup}{87}
\author{\bf Jianshu Zhang\textsuperscript{$^{\heartsuit}$\thanks{These authors contribute to this work equally.}}, ~ Dongyu Yao$^{\spadesuit*}$,  ~ Renjie Pi$^{\heartsuit}$, ~ Paul Pu Liang$^{\vardiamond}$, ~ Yi R. (May) Fung$^{\heartsuit}$\\
$^{\heartsuit}$Hong Kong University of Science and Technology \\ $^{\spadesuit}$Carnegie Mellon University
~~~~~~$^{\vardiamond}$Massachusetts Institute of Technology\\
\texttt{jianshu.zhang777@gmail.com}
~~~\texttt{raindy@cmu.edu} ~~~\texttt{rpi@ust.hk} \\
\texttt{ppliang@mit.edu} ~~~\texttt{yrfung@ust.hk}
}

\begin{document}
\maketitle

\begin{abstract}
    Visually linking matching cues is a crucial ability in daily life, such as identifying the same person in multiple photos based on their cues, even without knowing who they are. Despite the extensive knowledge that vision-language models (VLMs) possess, it remains largely unexplored whether they are capable of performing this fundamental task. To address this, we introduce \textbf{VLM2-Bench}, a benchmark designed to assess whether \textbf{VLM}s can \textbf{V}isually \textbf{L}ink \textbf{M}atching cues, with 9 subtasks and over 3,000 test cases. 
    Comprehensive evaluation across twelve VLMs, along with further analysis of various language-side and vision-side prompting methods, leads to a total of eight key findings. We identify critical challenges in models' ability to link visual cues, highlighting a significant performance gap. Based on these insights, we advocate for (i) enhancing core visual capabilities to improve adaptability and reduce reliance on prior knowledge, (ii) establishing clearer principles for integrating language-based reasoning in vision-centric tasks to prevent unnecessary biases, and (iii) shifting vision-text training paradigms toward fostering models' ability to independently structure and infer relationships among visual cues.\footnote{Project page: \url{https://vlm2-bench.github.io/}. \newline $^\spadesuit$Work was done while student was an intern at HKUST.}

\end{abstract}
\begin{figure}[t]
  \centering
  \includegraphics[width=0.5\textwidth]{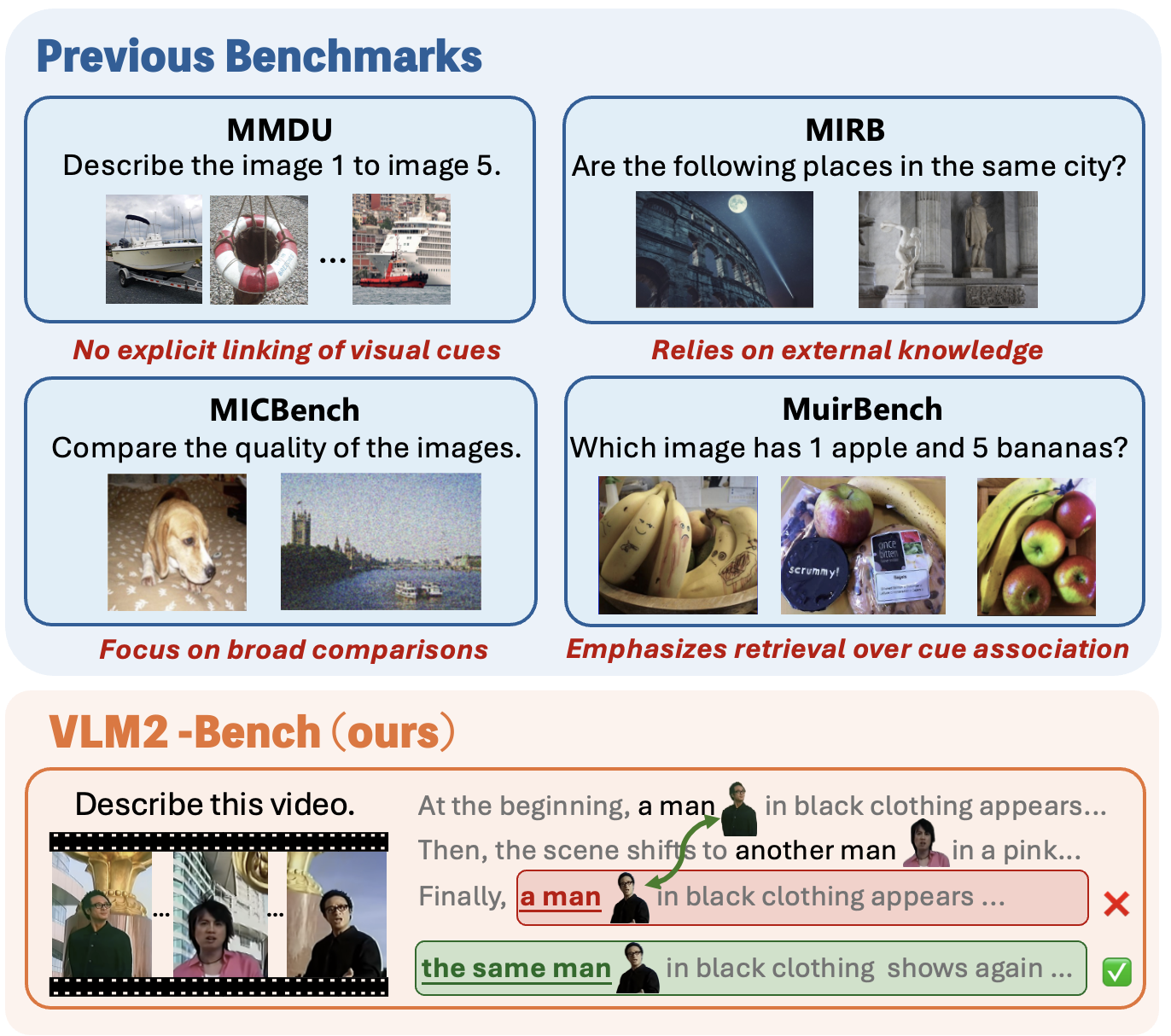} 
  \caption{\textcolor[HTML]{3d5d92}{\textbf{Previous benchmarks}} fail to assess the ability to link matching visual cues, whereas our \textcolor[HTML]{cd7d4e}{\textbf{VLM2-Bench}} explicitly tests this ability, as shown in the example where the model need to identify the reappearance of the same person by linking visual cues, like facial features or clothing, across non-adjacent frames. 
  }
  \label{fig:teaser}
  \vspace{-0.8em}
\end{figure}
\section{Introduction}
\begin{figure*}[t]
  \centering
  \includegraphics[width=0.99\textwidth]{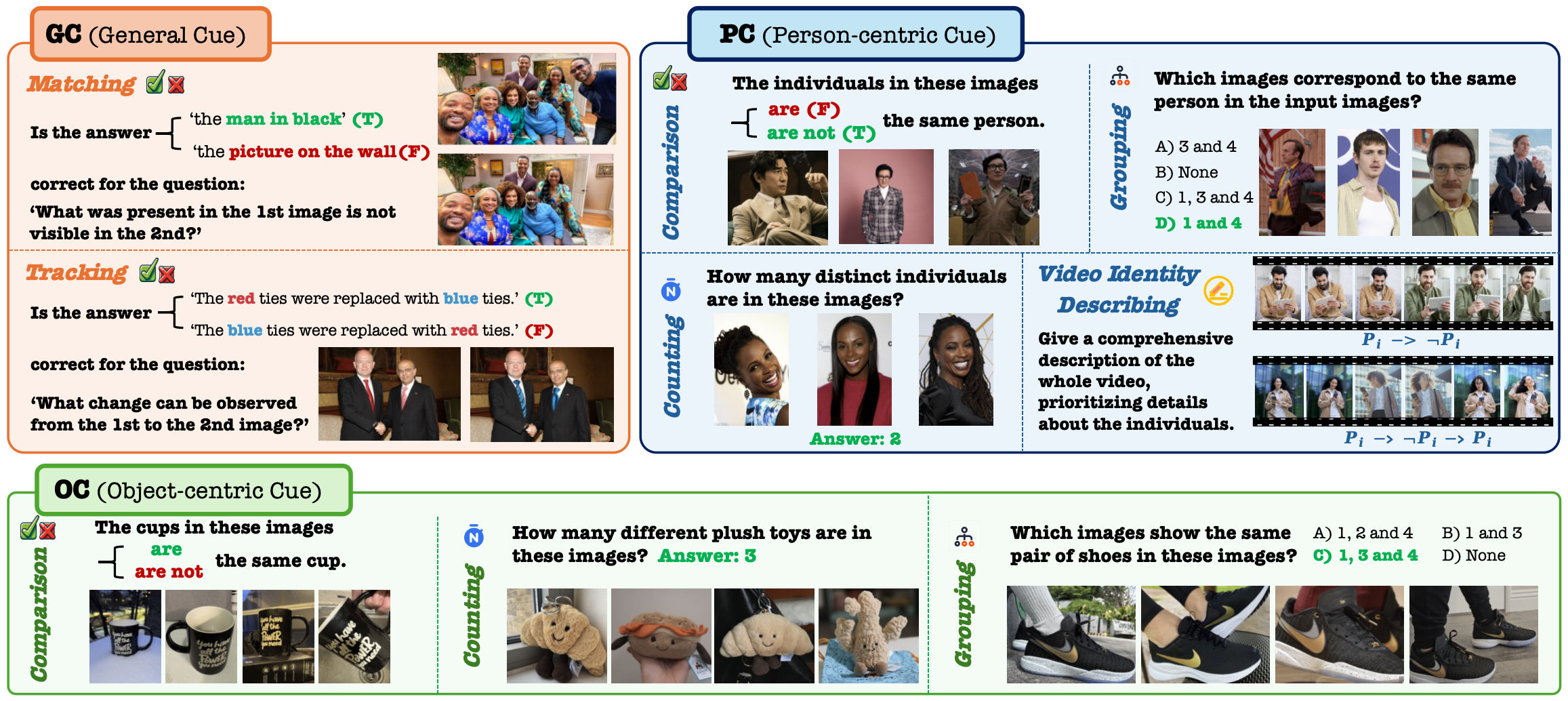} 
  \caption{Overview of \textbf{VLM2-Bench}. The benchmark is categorized into three subsets based on visual cues: GC (General Cue), OC (Object-centric Cue), and PC (Person-centric Cue), each comprising multiple subtasks. To comprehensively evaluate VLMs' ability to visually link matching cues, the benchmark includes diverse question formats—T/F \raisebox{-0cm}{\includegraphics[width=0.5cm]{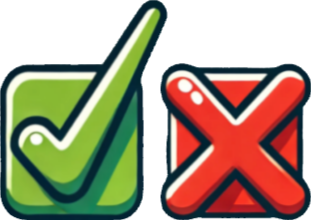}}, multiple-choice \raisebox{-0.1cm}{\includegraphics[width=0.4cm]{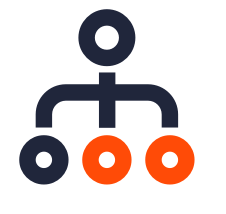}}, numerical \raisebox{-0.05cm}{\includegraphics[width=0.3cm]{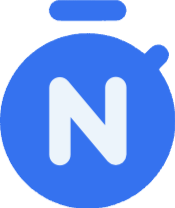}}, and open-ended \raisebox{0.0cm}{\includegraphics[width=0.3cm]{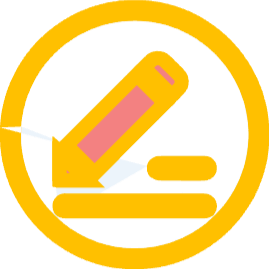}}—ensuring a comprehensive evaluation.}
  \label{fig:overview}
  \vspace{-0.6em}
\end{figure*}
Humans constantly link matching visual cues to navigate and understand their environment. For instance, we can determine whether objects, and individuals are the same simply by comparing their distinguishing visual features~\citep{face1,face2,object1}. This ability, often without needing additional background knowledge, is fundamental in our daily interactions with the world around us. However, while current vision-language models (VLMs)~\citep{internvl,llava-onevision,llava-video,Qwen2.5-VL} have demonstrated extensive knowledge and expanded their capabilities from single-image understanding to handling multiple images and videos, \textit{whether thay can effectively link matching visual cues across images or frames—an essential skill for coherent multimodal reasoning—remains an open question.} 

As shown in Figure~\ref{fig:teaser}, existing benchmarks on multiple images and videos
fall short in exploring this fundamental ability as they: (a) do not require explicitly linking visual cues across images or frames~\citep{liu2024mmdu, yu2019activitynet}; (b) rely on external knowledge rather than assessing models' ability to link explicitly visual cues~\citep{MIRB, liu2024mibench}; (c) emphasize broad and abstract visual comparisons rather than specific cue matching~\citep{MICBench, liu2024tempcompass}; and (d) focus on retrieval-based tasks rather than evaluating the direct association of visual cues across different visual contexts~\citep{wang2024muirbench}.

To bridge this gap, we introduce \textbf{VLM2-Bench}, a benchmark specifically designed to evaluate how well VLMs visually link matching cues. VLM2-Bench is structured around three types of visual cue connection: \textit{general cue}, \textit{person-centric cue}, and \textit{object-centric cue}, encompassing a total of eight subtasks. To balance scalability and quality, we design a semi-automated pipeline with human verification for further refinement. Additionally, our subtasks cover a variety of QA formats—including T/F, multi-choice, numerical, and open-ended questions—totaling over 3,000 question-answer pairs. To better evaluate model performance, we also design specific metrics tailored to various tasks.

We conduct a comprehensive evaluation of 8 open-source models and 3 commercial models on our VLM2-Bench. Despite VLMs generally possessing extensive knowledge, some models perform on par with, or even worse than, the chance-level baseline on our vision-centric tasks. Notably, even the most advanced commercial models fall short of human-level accuracy by over 30\%. This highlights the significant room for improvement in VLMs' ability to link visual cues. Furthermore, we introduce various language-side and vision-side prompting techniques to explore whether they can enhance the models' performance on the benchmark. Through experimental results and case studies, we present \textit{eight key observations}, hoping that these insights will guide future improvements in VLMs for vision-centric tasks.

\section{VLM2-Bench}
As shown in Figure~\ref{fig:overview}, VLM2-Bench is a benchmark designed to assess models' ability to visually link matching cues when processing multiple images or videos. This section introduces the three main categories of VLM2-Bench—\textit{general cue} (\S\ref{gc}), \textit{object-centric cue} (\S\ref{oc}), and \textit{person-centric cue} (\S\ref{pc})—detailing their associated subtasks, data collection process, and QA pair construction.

\begin{figure*}[t]
  \centering
  \includegraphics[width=0.99\textwidth]{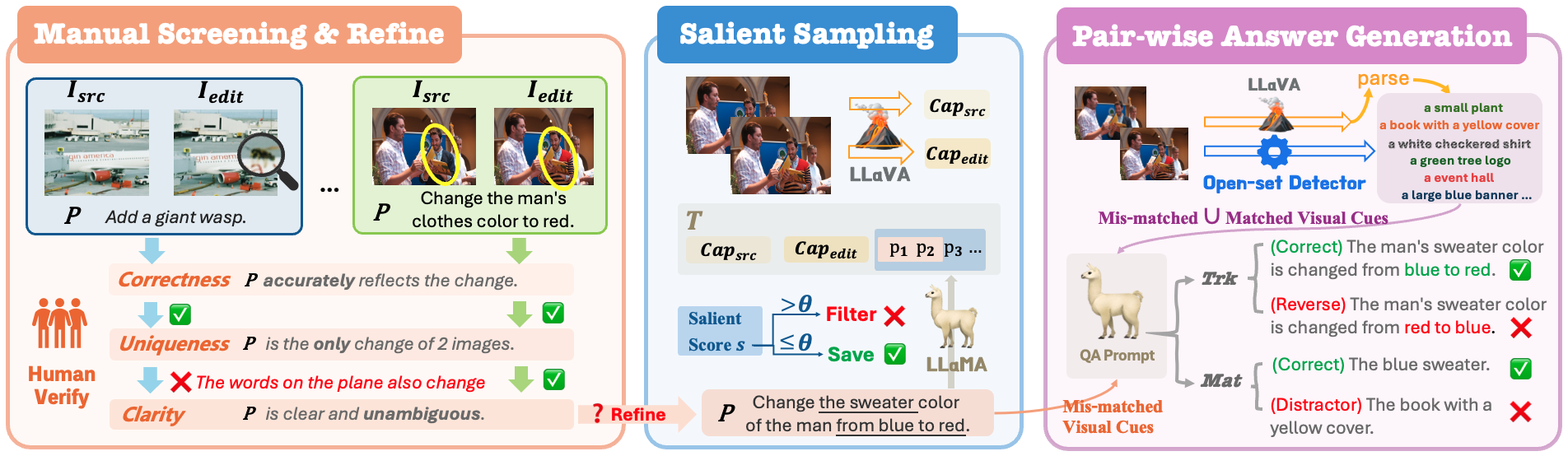} 
  \caption{Construction of \textbf{GC}: (i) We start by manually verifying the edited image data based on three key criteria. (ii) A VLM is then prompted to generate captions for each image, followed by salient score-based filtering to retain the challenging cases. (iii) Finally, visual cues are extracted from two sources and incorporated into a QA prompt, guiding an LLM to generate both positive and negative answer pairs. 
  }
  \label{fig:gc construction}
  \vspace{-0.6em}
\end{figure*}

\subsection{General Cue (GC)}
\label{gc}
GC is designed to assess a model's ability to link matching cues across diverse contexts, encompassing a broad range of \textit{general cues}. Given two images containing both matched and mismatched cues, an ideal model should accurately identify mismatched ones and associate matched ones.

\paragraph{Subtasks.}
Here we introduce two subtasks: (i) \textbf{\textit{Matching (Mat)}} evaluates a model’s ability to link corresponding visual cues across two images to determine whether they match. Instead of merely identifying differences, the model must associate identical visual elements in both images to recognize what has remained the same and what has changed. 
(ii) \textbf{\textit{Tracking (Trk)}} focuses on a model’s ability to track a specific visual cue that appears in only one of the two images and determine how it has changed. Rather than simply detecting a difference, the model must link the cue across contexts to understand the transformation process.

\paragraph{Data Collection.} 
We repurpose data from two image editing datasets~\citep{wei2024omniedit,ku2023imagenhub}, where each data sample includes an original image $I_{ori}$, an edited image with subtle modifications $I_{edit}$, and a corresponding edit instruction $\mathcal{P}$ describing the changes. Our data collection is carried out across two dimensions. First, to ensure diversity in the mismatched cues, GC encompasses various types of changes, such as instance-level modifications (e.g., add/remove, swap, attribute change), which focus on specific items, as well as environment-level changes.

\paragraph{QA Construction.}
We predefine a T/F question template for \textit{Mat} and \textit{Trk} with a placeholder for the candidate answer (refer to Appendix~\ref{appendix: more details on benchmark construction}). Figure~\ref{fig:gc construction} illustrates the construction process, which follows a three-stage approach. 

\textit{Manual Screening \& Refinement:} We ensure that $\mathcal{P}$ accurately reflects the changes (correctness), corresponds uniquely to the modified cues (uniqueness), and is unambiguous (clarity).

\textit{Salient Sampling:} Here, we automate the removal of overly simple cases (e.g., mismatched cues are too salient). To achieve this, a VLM first generates separate descriptions for \( I_{ori} \) and \( I_{edit} \), denoted as \( Cap_{ori} \) and \( Cap_{edit} \). These descriptions are then combined with \( \mathcal{P} \) into a single passage using a predefined template \(\mathcal{T}\) (see Table \ref{template salient score} for details). The probability assigned by a language model (e.g., Llama3-8B~\citep{dubey2024llama}) to \( \mathcal{P} \) given this text-based information is used to compute the salient score, formulated as:

\begin{equation}
S_{\text{salient}} = \frac{1}{|\mathcal{P}|} \sum_{i=1}^{|\mathcal{P}|} \log P_{\theta}(p_i \mid C \cup p_{<i}),
\end{equation}

\noindent where \( \mathcal{P} = \{p_1, p_2, ..., p_{|\mathcal{P}|}\} \) represents the tokenized \(\mathcal{P}\), and \( C = \mathcal{T} (Cap_{ori}, Cap_{edit}) \) denotes the context filled with template \(\mathcal{T}\). Samples with scores below \( \theta \) (-2.0 here) are retained, ensuring that the benchmark includes more challenging examples requiring nuanced visual cue association. 

\textit{Pair-wise Answer Generation:} Finally, we extract visual cues using a dual-level approach. First, cues parsed from VLM-generated descriptions compensate for the limitations of open-set detectors when handling out-of-distribution scenes. Meanwhile, the open-set detector~\citep{wu2022grit} extracts fine-grained cues that VLMs might overlook. With these extracted cues, we prompt an LLM to generate a pair of answers for \textit{Mat} and \textit{Trk}, each consisting of one positive and one negative answer.

\subsection{Object-centric Cue (OC)}
\label{oc}
OC aims to assess a model's ability to link matching cues associated with everyday objects using \textit{object-centric cues}. Even when encountering an object for the first time, a well-aligned model should be able to leverage its unique visual cues to establish associations, enabling it to recognize and track the object across different scenes. This capability is essential for coherent perception and interaction in real-world deployments.

\paragraph{Subtasks.} 
\label{subtasks}
Based on the complexity of linking cues to solve the problem, we define three subtasks in OC. (i) \textbf{\textit{Comparison (Cpr)}} requires the model to determine whether the objects appearing in different images are the same. This task primarily assesses the model’s ability to perceive visual consistency or change. Notably, we observe that models exhibit significant model-specific bias when making a binary decision~\citep{pair, ye2024justice, song2024large,li2024naturalbench}, leading to discrepancies between results and their actual capabilities. To mitigate this, we introduce consistency-pair validation, where for each statement (e.g., ``X is Y”, with the answer being T), we generate a corresponding negation (e.g., ``X is not Y”, with the answer being F). The model is only considered correct if it correctly answers both statements, ensuring consistency in its decision-making.
(ii) \textbf{\textit{Counting (Cnt)}} involves identifying the number of unique objects, requiring the model not only to recognize variations or consistencies but also to track distinct cues to avoid double-counting the same object. (iii) \textbf{\textit{Grouping (Grp)}}, the most challenging one, requires the model to identify all instances of the same object, building on precise cue matching across multiple images.

\paragraph{Data Collection.}  
We manually collect various categories of everyday objects (e.g., pets, cups) from multiple online resource~\footnote{https://www.amazon.com/, https://lens.google/, and https://jellycat.com/.}. For each category, we define multiple subcategories and collect a set of images \( \mathcal{I}_{O_i}\)—four images that depict the same object in different scenarios. Additionally, we also collect a set \( \mathcal{I}_{\neg O_i} \), consisting of four images of different objects, each containing some matching visual cues with \( \mathcal{I}_{O_i} \), which are used as distractors.

\paragraph{QA Construction.}
For each subtask, we define a question template that includes a placeholder for \( \mathcal{I}_{O_i} \), which allows us to tailor the question based on different objects (see Appendix~\ref{appendix: more details on benchmark construction}). For answer generation, we first curate the multi-image sequences according to predefined rules. For each specific sequence, we generate the ground truth answers for the questions related to \textit{Cpr}, \textit{Cnt}, and \textit{Grp}.

\subsection{Person-centric Cue (PC)}
\label{pc}
PC aims to evaluate a model's ability to link \textit{person-centric cues}. While a model cannot memorize every individual, it should possess the capability to associate the same person across different images or frames by leveraging distinctive visual cues such as facial features, clothing, or body posture. This ability is essential for ensuring coherent perception of human actions and is a fundamental requirement for real-world VLM applications.

\paragraph{Subtasks.} 
Similar to OC's subtasks (refer to \S\ref{subtasks}), PC includes (i) \textbf{\textit{Comparison (Cpr)}}, (ii) \textbf{\textit{Counting (Cnt)}}, and (iii) \textbf{\textit{Grouping (Grp)}}. However, unlike objects, individuals can be observed through their actions in videos. Therefore, we introduce (iv) \textbf{\textit{Video Identity Describing (VID)}}. This subtask assesses whether a model can correctly link the same person by analyzing its description of a video containing that person.

\paragraph{Data Collection.}  
We manually select several individuals, each denoted as \( \mathcal{P}_i \). For each individual, we collect \( \mathcal{I}_{\mathcal{P}_i}\)—4 images depicting the same individual. For each image \( I_i  \in  \mathcal{I}_{\mathcal{P}_i} \), we select the distractor images \( I_{\neg i} \notin \mathcal{I}_{\mathcal{P}_i} \) that has the highest CLIP similarity~\citep{hessel2021clipscore}. This allows us to obtain images of different individuals where most cues are matched.
For the subtask of \textit{VID}, we collect videos of different individuals, denoted as \( V_{\mathcal{P}_i} \), and pair each with another video \( V_{\neg \mathcal{P}_i} \) featuring a different individual with highly similar cues (e.g., actions, scene, clothing). We then construct two video sequences: 
(i) \( \mathcal{P}_i \xrightarrow{} \neg \mathcal{P}_i \), assessing the model's ability to distinguish individuals. 
(ii) \( \mathcal{P}_i \xrightarrow{} \neg \mathcal{P}_i \xrightarrow{} \mathcal{P}_i \), evaluating whether the model detects changes and links the final occurrence of \( \mathcal{P}_i \) to its first appearance.


\paragraph{QA Construction.} The construction for the overall QA in PC’s \textit{Cpr}, \textit{Cnt}, and \textit{Grp} subtasks follows a similar approach to OC. For the \textit{VID} task, we emphasize the model's ability to describe individuals when designing open-ended questions, aiming to better test the model's capacity to link individuals appearing in different scenes.

\begin{figure}[t]
  \centering
  \includegraphics[width=0.49\textwidth]{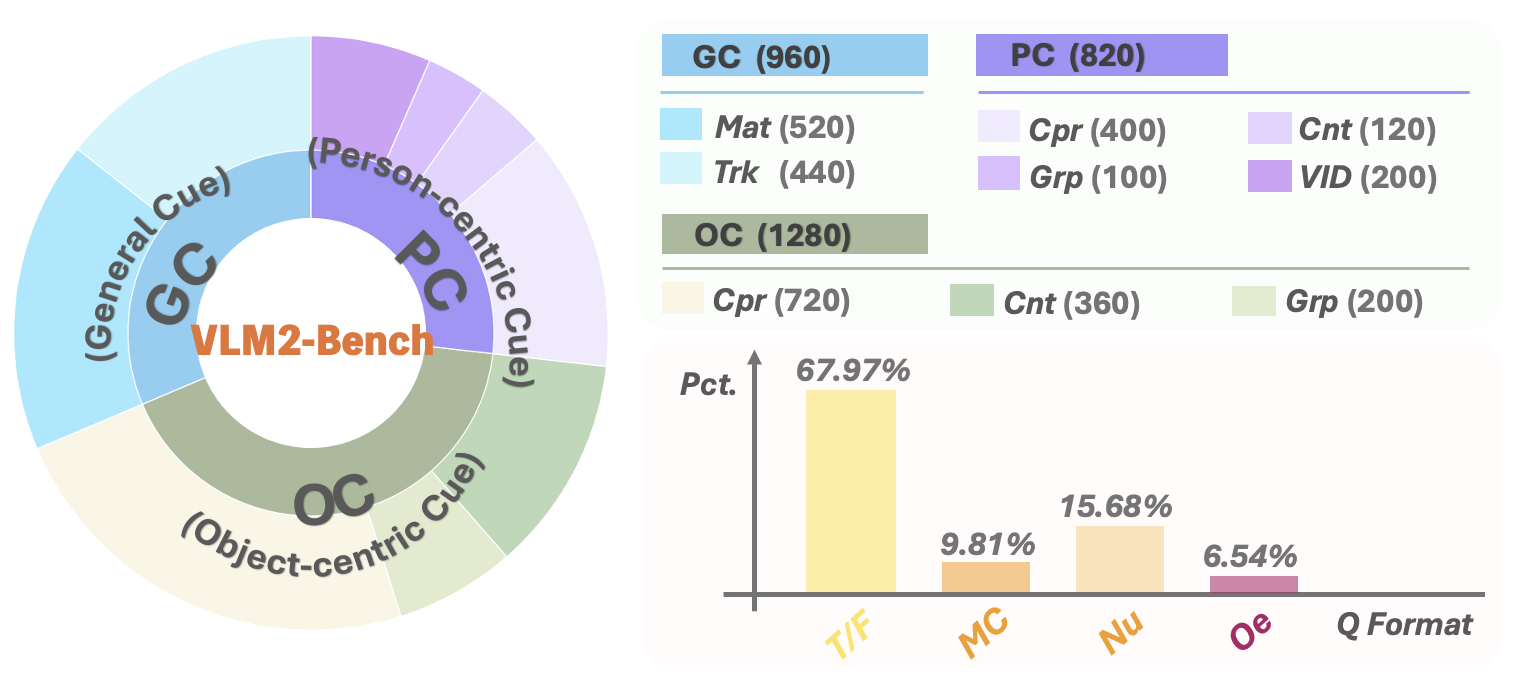} 
  \caption{Statistical overview of \textbf{VLM2-Bench}. The pie chart shows the distribution of 9 subtasks across the 3 main categories of visual cues. The bar plot illustrates the percentage breakdown by question format. 
  }
  \label{fig:bench statistics main}
\end{figure}

\begin{figure*}[t]
    \captionsetup{type=table}
    \centering
    \begin{minipage}{0.99\textwidth}
        \centering
        \resizebox{1\textwidth}{!}{%
        \begin{tabular}{l||cc|ccc|cccc|cc}
            \toprule
            \textbf{Baselines or Models} & \multicolumn{2}{c|}{\textbf{GC}} & \multicolumn{3}{c|}{\textbf{OC}} & \multicolumn{4}{c|}{\textbf{PC}} & \multicolumn{2}{c}{\textbf{Overall*}}\\
            & \textit{Mat} & \textit{Trk} & \textit{Cpr} & \textit{Cnt} & \textit{Grp} & \textit{Cpr} & \textit{Cnt} & \textit{Grp} & \textit{VID} & Avg & \textbf{$\Delta_{human}$} \\
            \midrule
            \textcolor{black!50}{Chance-Level} & 
            \textcolor{black!50}{25.00} & 
            \textcolor{black!50}{25.00} & 
            \textcolor{black!50}{50.00} & 
            \textcolor{black!50}{34.88} & 
            \textcolor{black!50}{25.00} & 
            \textcolor{black!50}{50.00} & 
            \textcolor{black!50}{34.87} & 
            \textcolor{black!50}{25.00} & 
            \textcolor{black!50}{-} & 
            \textcolor{black!50}{33.72} & 
            \textcolor{black!50}{-61.44} \\
            Human-Level  & 95.06 & 98.11 & 96.02 & 94.23 & 91.00 & 97.08 & 92.87 & 91.17 & 100.00 & 94.44 & 0.00 \\
            \midrule
            LLaVA-OneVision-7B    & 16.60 & 13.70 & 47.22 & 56.17 & 27.50 & 62.00 & 46.67 & 37.00 & 47.25 & 38.36 & -56.08 \\
            LLaVA-Video-7B        & 18.53 & 12.79 & 54.72 & 62.47 & 28.50 & 62.00 & 66.91 & 25.00 & 59.00 & 41.37 &  -53.07 \\
            LongVA-7B             & 14.29 & 19.18 & 26.67 & 42.53 & 18.50 & 21.50 & 38.90 & 18.00 & 3.75  & 24.95 & -69.49 \\
            mPLUG-Owl3-7B         & 17.37 & 18.26 & 49.17 & 62.97 & 31.00 & \cellcolor{yellow!15}63.50 & 58.86 & 26.00 & 13.50 & 40.89 & -53.55 \\
            Qwen2-VL-7B  & 27.80 & 19.18 & 68.06 & 45.99 & 35.00 & 61.50 & 58.59 & 49.00 & 16.25 & 45.64 & -48.80 \\
            Qwen2.5-VL-7B & \cellcolor{yellow!45}35.91 & \cellcolor{yellow!75}43.38 & 71.39 & 41.72 & 47.50 & \cellcolor{yellow!75}80.00 & 57.98 & \cellcolor{yellow!75}69.00 & 46.50 & 55.86 & -38.58  \\
            InternVL2.5-8B        & 21.24 & 26.03 & 53.33 & 55.23 & 46.50 & 51.50 & 60.00 & \cellcolor{yellow!15}52.00 & 5.25  & 45.73 & -48.71 \\
            InternVL2.5-26B        & 30.50 & 30.59 & 43.33 & 51.48 & \cellcolor{yellow!15} 52.50 & 59.50 & 59.70 & \cellcolor{yellow!45}61.00 & 21.75 & 48.58 & -45.86 \\
            \midrule
            Gemini-2.0-flash & 1.54 & 14.61 & 51.67 & 35.57 & 23.00 & 49.00 & 30.24 & 21.00 & - & 28.33 & -66.11 \\
            Claude-3.7-sonnet          & \cellcolor{yellow!15}33.72 & \cellcolor{yellow!15}36.41 & \cellcolor{yellow!45}74.44 & \cellcolor{yellow!15}73.02 & \cellcolor{yellow!75}64.50 & \cellcolor{yellow!45}67.50 & \cellcolor{yellow!15}67.00 & \cellcolor{yellow!15} 60.00 & \cellcolor{yellow!15}61.25 & \cellcolor{yellow!45}59.57 & -34.87 \\
            GPT-4o-2024-08-06          & \cellcolor{yellow!75}37.45 & \cellcolor{yellow!45}39.27 & \cellcolor{yellow!15}74.17 & \cellcolor{yellow!75}80.62 & \cellcolor{yellow!45}57.50 & 50.00 & \cellcolor{yellow!75}90.50 & 47.00 & \cellcolor{yellow!45}66.75 & \cellcolor{yellow!75}59.56 & -34.88 \\
            GPT-4o-2024-11-20         & 18.53 & 29.68 & \cellcolor{yellow!75}81.67 & \cellcolor{yellow!45}77.08 & \cellcolor{yellow!45}57.50
 & 56.00
 & \cellcolor{yellow!45}78.39 & 47.00 & \cellcolor{yellow!75}76.55 & \cellcolor{yellow!15}55.73  & -38.71 \\
            \bottomrule
        \end{tabular}
        }
    \end{minipage}
    \caption{Evaluation results on \textbf{VLM2-Bench}, covering \textit{Mat} (Matching), \textit{Trk} (Tracking), \textit{Cpr} (Comparison), \textit{Cnt} (Counting), \textit{Grp} (Grouping), and \textit{VID} (Video Identity Describing). The \colorbox{yellow!75}{highest}, \colorbox{yellow!45}{second}, and \colorbox{yellow!15}{third} highest scores are highlighted. *: Overall excludes the \textit{VID} due to the lack of a chance-level baseline for open-ended tasks.}
    \label{exp:main_exp}
\end{figure*}

\subsection{Benchmark Statistics}
\label{bench_statistics_main}
Our benchmark is organized into three main categories, comprising a total of 9 subtasks. After careful verification, it contains 3,060 question-answer pairs, with varying formats including T/F, multi-choice (MC), numerical (Nu), and open-ended (Oe). To ensure the quality of the annotations, we perform an inter-annotator agreement (IAA) evaluation~\citep{thorne2018fever} involving three annotators, resulting in a high Fleiss' Kappa score~\citep{fleiss1971measuring} of 0.983. Figure~\ref{fig:bench statistics main} presents the distribution of these subtasks across the three categories, along with the breakdown of different question formats. For additional details, refer to Appendix~\ref{appendix: statistics}.


\section{Evaluation }
\subsection{Metric Design}
\label{metrics}

\paragraph{T/F}  \textit{(Matching, Tracking, Comparison)}: Accuracy is computed based on paired evaluation, where a response is correct only if it answers \(T\) (ground-truth True) and \(F\) (ground-truth False) correctly. The overall accuracy across \(N\) test pairs is:

\begin{equation}
    Acc_{pair} = \frac{\sum_{i=1}^{N} \left( T_i^+ \cap F_i^- \right)}{N},
\end{equation}

\noindent where \( T^+ \) and \( F^- \) denote correct predictions for \(T\) and \(F\), respectively.






\paragraph{Numerical} \textit{(Counting)}: Absolute matching alone does not effectively reflect the severity of errors in numerical responses. To measure the extent of the error between the predicted count \( \hat{N}_i \) and ground truth \( N_i \), we introduce \(Acc_{num}\). The first step is to calculate the normalized error:

\begin{equation}
\epsilon_i = \frac{\left| \hat{N}_i - N_{i} \right|}{\max\left(N_{i} - 1, N^{img}_i - N_i\right)},
\end{equation}

\noindent where \( N^{img}_i \) is the number of input images. We define \( w_i = \max (\{N^{\text{img}}_i\}_{i=1}^n )/{N^{\text{img}}_i} \) to penalize errors in cases with fewer images and introduce \( \alpha \) as an error amplification factor. The final accuracy over \( n \) cases is:

\begin{equation}
Acc_{num} = 1 - \frac{1}{n} \sum_{i=1}^{n} w_i \cdot \epsilon_i^{\alpha}.
\end{equation}


\paragraph{Multi-choice} \textit{(Grouping)}: Accuracy is the proportion of correctly predicted choices.

\paragraph{Open-ended} \textit{(Video Identity Describing)}: We use GPT-4o to score model's descriptions, in combination with rule-based scoring prompts. The final accuracy \(Acc_{oe}\) is obtained by averaging the scores of all open-ended responses and rescaling them to the range of [0,1]. Additionally, we perform manual verification of GPT-4o’s scoring. For each model, we randomly sample 20 scored responses for review, and find only 2 instances with discrepancies, resulting in an accuracy rate of 98.89\% (178/180). Refer to Appendix~\ref{appendix: prompting approaches} for more details.

\subsection{Evaluation Setup}
\paragraph{Evaluated Models.} We evaluate eight open-source VLMs that support multiple-image or video input: LLaVA-OneVision~\citep{llava-onevision}, LLaVA-Video~\citep{llava-video}, LongVA~\citep{longva}, mPLUG-Owl3~\citep{mplug-owl3}, Qwen2-VL~\citep{Qwen2-VL}, Qwen2.5-VL~\citep{Qwen2.5-VL}, and InternVL2.5~\citep{internvl}. Additionally, we include the commercial models GPT-4o~\citep{gpt4o}, Claude-3.7-sonnet, and Gemini-2.0-flash for comparison for comparison.

\paragraph{Baselines.} We introduce chance-level and human-level baselines (details are in Appendix~\ref{appendix: baselines}).

\subsection{Results and Findings}
\paragraph{Results.} Table~\ref{exp:main_exp} presents the comprehensive performance of various models across the  three categories -- General Cue (GC), Object-centric Cue (OC), and Person-centric Cue (PC) -- of our VLM2-Bench, covering a total of nine subtasks.

\paragraph{Finding I: Simple tasks for humans pose significant challenges for VLMs.} We observe that humans achieve near-perfect accuracy across most tasks in our VLM2-Bench. In contrast, even state-of-the-art closed-source models perform significantly lower than humans. For open-source models, many show performance comparable to the chance-level baseline or only slightly outperform it. Specifically, for the \textit{VID}, humans can easily achieve 100\% accuracy in distinguishing and linking individuals in a video. Errors mainly arise from failing to recognize individuals after changes or misidentifying reappearing persons as new.

\begin{table}[t]
\centering
\resizebox{0.49\textwidth}{!}{%
\begin{tabular}{l||cccc||cccc}
    \toprule
    Model&\multicolumn{4}{c}{Matching (\textit{Mat})}&\multicolumn{4}{c}{Tracking (\textit{Trk})}\\ \hline
     & \textbf{A/R} & \textbf{Swp} & \textbf{Attr} & \textbf{Env} & \textbf{A/R} & \textbf{Swp} & \textbf{Attr} & \textbf{Env} \\ \hline
    LV-OV & \cellcolor[HTML]{d9eaf4}50.68 & \cellcolor[HTML]{fbecd9}49.15 & 53.45 & 52.50  & \cellcolor[HTML]{fbecd9}27.27 & 	\cellcolor[HTML]{d9eaf4}45.51	 & 57.50	 & 70.59\\ \hline
    LV-Vid & 56.08 & \cellcolor[HTML]{fbecd9}49.15 & 53.45 & \cellcolor[HTML]{d9eaf4}51.25  & \cellcolor[HTML]{fbecd9}46.75	 & \cellcolor[HTML]{d9eaf4}48.88 & 	52.50 & 	67.65\\ \hline
    LongVA & \cellcolor[HTML]{fbecd9}37.84 & 46.58 & 53.45 & \cellcolor[HTML]{d9eaf4}46.25  & \cellcolor[HTML]{d9eaf4}46.10	 & 49.44	 & \cellcolor[HTML]{fbecd9}42.50 & 	60.29\\ \hline
    Owl3 & 54.73 & \cellcolor[HTML]{d9eaf4}52.56 & 55.17 & \cellcolor[HTML]{fbecd9}50.00  & \cellcolor[HTML]{fbecd9}41.56	 & \cellcolor[HTML]{d9eaf4}48.88	 & 55.00	 & 73.53\\ \hline
    Qw2-VL & \cellcolor[HTML]{d9eaf4}53.68 & \cellcolor[HTML]{fbecd9}52.56 & 55.17 & 68.75  & 65.58 & 	\cellcolor[HTML]{fbecd9}62.90 & 	77.50 & 	\cellcolor[HTML]{d9eaf4}63.93\\ \hline
    Qw2.5-VL & \cellcolor[HTML]{d9eaf4}64.19 & \cellcolor[HTML]{fbecd9}55.62 & 74.14 & 67.50  & \cellcolor[HTML]{d9eaf4}61.69 & 	69.10 & 	\cellcolor[HTML]{fbecd9}55.00 & 	64.71\\ \hline
    In2.5-8B & 64.86 & \cellcolor[HTML]{fbecd9}51.28 & \cellcolor[HTML]{d9eaf4}52.07 & 66.25 &  \cellcolor[HTML]{fbecd9}54.55 & 	67.42 & 	62.50 & 	\cellcolor[HTML]{d9eaf4}60.65\\ \hline
    In2.5-26B & 60.81 & \cellcolor[HTML]{fbecd9}51.71 & \cellcolor[HTML]{d9eaf4}58.62 & 61.25 &  \cellcolor[HTML]{d9eaf4}56.49 & 	62.92	 & \cellcolor[HTML]{fbecd9}47.50 & 	66.18\\ \hline
    GPT-4o & 75.00 & \cellcolor[HTML]{d9eaf4}61.97 & \cellcolor[HTML]{fbecd9}56.90 & 70.00 &  68.83 & 	\cellcolor[HTML]{d9eaf4}67.98	 & 67.50 & 	\cellcolor[HTML]{fbecd9}64.71\\ 
    \bottomrule
    \end{tabular}
}   
    \caption{Breakdown of four mis-matched cue types in two subtasks of GC. For each model, the \colorbox[HTML]{fbecd9}{highest} and \colorbox[HTML]{d9eaf4}{second highest}  error (\%) per subtask are highlighted.}
    \label{exp: breakdown gc}
\end{table}

\paragraph{Finding II: Relatively consistent error patterns in \textit{Mat} and \textit{Trk} of GC.} Table~\ref{exp: breakdown gc} shows that models struggle with mismatched cues due to swap in \textit{Mat}, which requires linking two completely different cues. To identify what has changed, models must first link and match all the other cues in the context before they can determine that the swapped cue has been transformed. 
This task requires a deeper understanding of how cues relate to each other across different instances. In contrast, \textit{Trk} challenges models with mismatched cues due to add/remove, which focuses on tracking how a specific cue changes. This suggests that when there is a cue that appears only once, the model struggles to link the non-appearing cue with the appearing cue to track the transformation process effectively. This limitation reveals models' difficulty in handling cases where certain cues are missing but still need to be linked to understand the dynamic changes. 

\paragraph{Finding III: Models perform better in linking person-centric cues than object-centric cues.} We selected the top three open-source models (Qwen2.5-VL-8B, InternVL2.5-8B, InternVL2.5-26B) and compared their performance on the three shared tasks (\textit{Cpr, Cnt, Grp}) in both OC and PC. Results show that, on average, the performance on PC is higher than on OC by 7.65\%, 9.75\%, and 11.83\% for the tasks of \textit{Cpr, Cnt, Grp}, respectively. This could be due to the fact that, during training on person-related data, models are likely provided with explicit person names as anchors to person-centric cues, which helps the models better distinguish different individuals. In contrast, objects are typically trained using general category names, which may not provide such clear distinctions. Additionally, these models might have been specifically trained on large datasets that emphasize differentiating and linking individuals~\citep{pi2024personalized,human-data}, thereby enhancing their ability to link person-centric cues.

\begin{table}[t]
\centering
\resizebox{0.50\textwidth}{!}{%
\begin{tabular}{c||cc|ccc|ccc}
\toprule
\textbf{Res.} & \multicolumn{2}{c|}{\textbf{GC}} & \multicolumn{3}{c|}{\textbf{OC}} & \multicolumn{3}{c}{\textbf{PC}} \\
 & \textit{Mat} & \textit{Trk} & \textit{Cpr} & \textit{Cnt} & \textit{Grp} & \textit{Cpr} & \textit{Cnt} & \textit{Grp} \\
\hline
\multicolumn{9}{c}{\textbf{Qwen2.5-VL-7B}} \\
\hline
Origin  & 35.91 & 43.38 & 71.39 & 41.72 & 47.50 & 80.00 & 57.98 & 69.00 \\
$\downarrow$ ×2 & 25.10 & 40.18 & 64.17 & 45.75 & 42.50 & 76.00 & 60.45 & 70.00 \\
$\downarrow$ ×4 & 19.69 & 33.33 & 52.78 & 42.25 & 33.00 & 64.50 & 57.15 & 61.00 \\
$\downarrow$ ×8 & 13.90 & 24.66 & 43.33 & 43.22 & 24.00 & 57.00 & 48.65 & 52.00 \\
$\downarrow$ ×16 & 9.27 & 18.72 & 34.17 & 38.86 & 22.50 & 45.50 & 47.01 & 41.00 \\
\hline
\multicolumn{9}{c}{\textbf{InternVL2.5-8B}} \\
\hline
Origin  & 21.24 & 26.03 & 53.33 & 55.23 & 47.50 & 51.50 & 60.00 & 52.00 \\
$\downarrow$ ×2 & 10.42 & 19.63 & 72.50 & 53.33 & 45.00 & 50.50 & 53.67 & 50.00 \\
$\downarrow$ ×4 & 11.97 & 20.09 & 69.72 & 52.77 & 47.00 & 51.00 & 54.99 & 49.00 \\
$\downarrow$ ×8 & 10.04 & 16.89 & 68.33 & 49.96 & 45.00 & 52.00 & 52.86 & 49.00 \\
$\downarrow$ ×16 & 3.47 & 14.16 & 61.39 & 43.30 & 47.50 & 49.00 & 51.06 & 50.00 \\
\bottomrule
\end{tabular}
}
\caption{Models' performance at the original resolution and with various compression levels. Results show a clear performance decline as image quality decreases, indicating that VLM2-Bench requires models to perceive and distinguish fine-grained visual details.}
\label{visual_sanity}
\end{table}

\subsection{Visual Bias Sanity Check}
To assess whether models genuinely rely on fine-grained visual cues—rather than shortcut biases such as global layout or coarse semantics, we conduct a sanity check via image resolution ablation. Specifically, we evaluate two models (Qwen2.5-VL-7B and InternVL2.5-8B) under different levels of image compression, reducing the resolution by factors of 2, 4, 8, and 16.

As shown in Table~\ref{visual_sanity},we observe a consistent performance drop as the image resolution decreases. This trend suggests that models do rely on detailed visual cues to perform well, rather than exploiting high-level layout or textual artifacts. These results highlight two key insights: (i) Our benchmark tasks indeed require models to perceive and distinguish fine-grained visual differences, rather than exploiting shallow biases. (ii) The performance8
sensitivity to visual degradation provides evidence that top-performing models are engaging in genuine visual understanding—which supports the benchmark's role in probing visual linking ability under realistic, perception-driven settings.

\begin{figure}[h]
  \centering
  \includegraphics[width=0.35\textwidth]{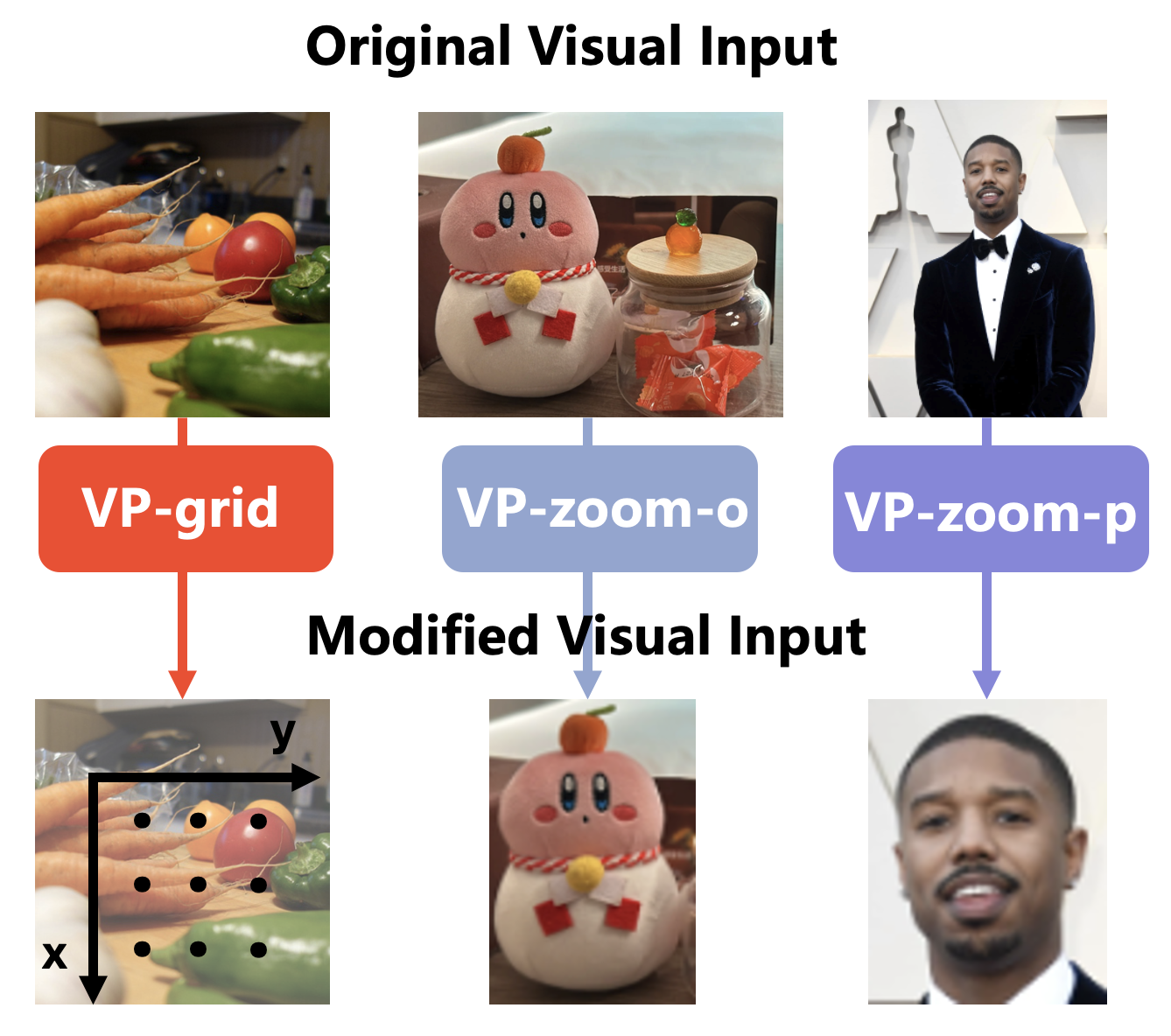} 
  \caption{Visualization of three visual prompting (VP) approaches we adopted in Section~\ref{sec: prompting}. From left to right: the VPs used for GC, OC, and PC, respectively.}
  \label{fig:vp}
\end{figure}

\section{How Prompting Methods Affect VLMs}
\label{sec: prompting}
In this section\footnote{Due to space limits, we reference most case studies, figures, and details in the Appendix within this section.}, we investigate various prompting methods (language-side and vision-side) to evaluate their impact on performance in VLM2-Bench. We select the top 3 performing open-source models (Qwen2.5-VL-8B, InternVL2.5-8B, InternVL2.5-26B), along with GPT-4o, and explore different approaches of CoT~\citep{cot1,cot2} and visual prompting (VP)~\citep{lei2024scaffoldingcoordinatespromotevisionlanguage, vp-zoom-in} (refer to Appendix~\ref{appendix: prompting approaches} for details). The goal is to investigate whether these techniques can improve performance across the benchmark and to identify the underlying factors that contribute to their success or failure. 

\subsection{Probing for General Cue (GC)}
\label{probing gc}
\paragraph{Methods.} (i) \textbf{CoT-normal} (Table~\ref{cot prompt}) encourages the model to solve the task step by step, allowing it to reason through the problem. (ii) \textbf{CoT-special} (Table~\ref{cot-special}) guides the model to solve the task using a thought process closer to how humans typically approach it. (iii) \textbf{VP-grid} (Figure~\ref{fig:vp-grid}) is adapted from previous work ~\citep{lei2024scaffoldingcoordinatespromotevisionlanguage} for our tasks, overlaying a dot matrix on the image as visual anchors to provide positional references and enhance the model's performance.

\begin{figure}[t]
  \centering
  \begin{subfigure}[b]{0.48\textwidth}
    \centering
    \includegraphics[width=\textwidth]{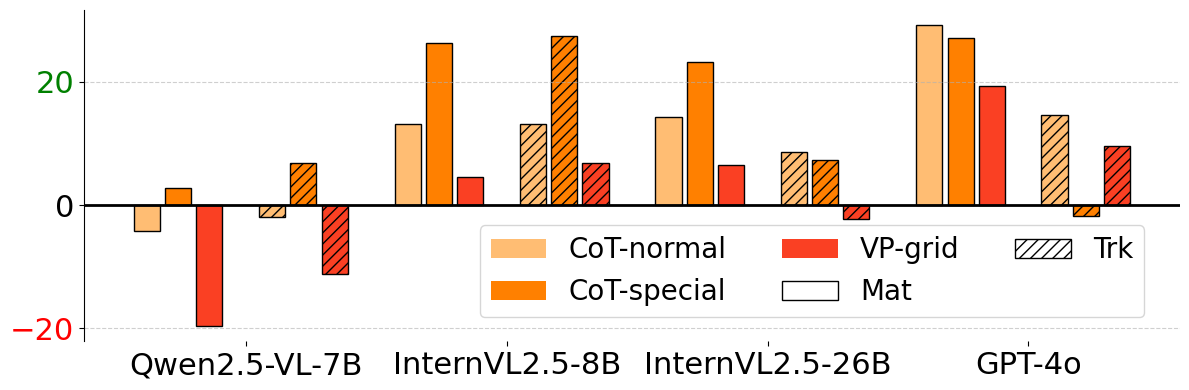}
    \caption{Results of CoT-normal, CoT-special, and VP-grid on GC.}
    \label{fig:gc-analysis}
  \end{subfigure}
  \\ 
  \begin{subfigure}[b]{0.48\textwidth}
    \centering
    \includegraphics[width=\textwidth]{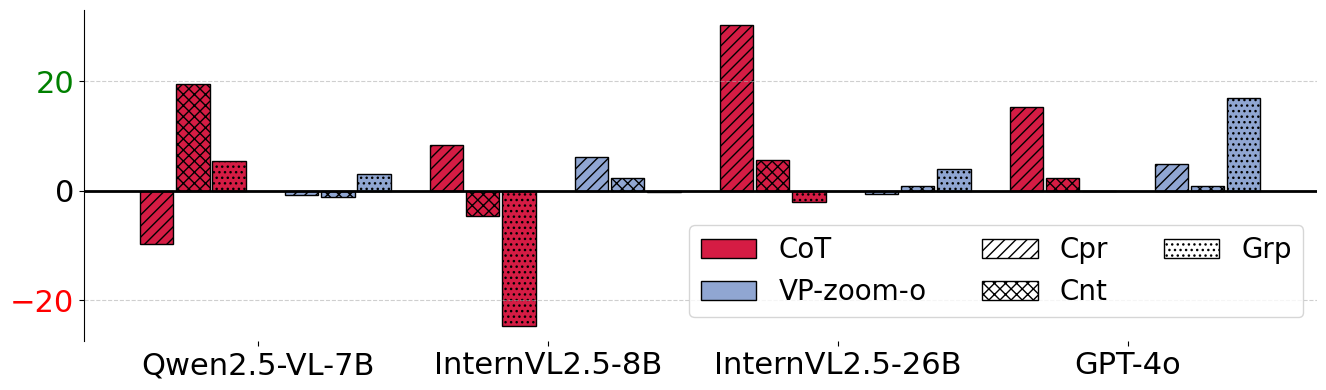}
    \caption{Results of CoT and VP-zoom-o on OC.}
    \label{fig:oc-analysis}
  \end{subfigure}
  \\ 
  \begin{subfigure}[b]{0.48\textwidth}
    \centering
    \includegraphics[width=\textwidth]{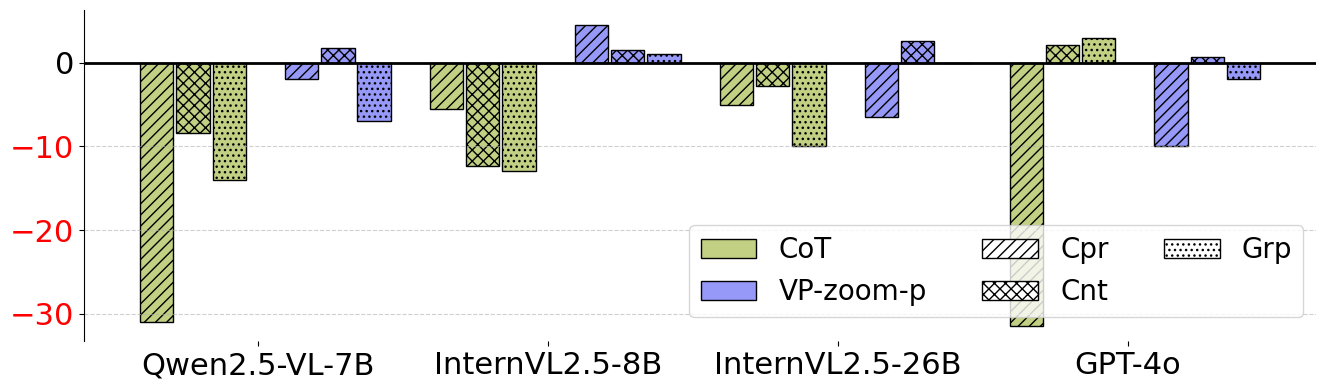}
    \caption{Results of CoT and VP-zoom-p on PC.}
    \label{fig:pc-analysis}
  \end{subfigure}
  \caption{Performance \textcolor{teal}{gains} and \textcolor{red}{losses} (\%) when applying different prompting methods on VLM2-Bench. (a) shows results on GC using CoT-normal, CoT-special, and VP-grid; (b) presents results on OC with CoT and VP-zoom-o; and (c) reports results on PC with CoT and VP-zoom-p. Detailed analyses are provided in Section~\ref{probing gc}, Section~\ref{probing oc}, and Section~\ref{probing pc}, respectively.}

  \label{fig:analysis}
\end{figure}

\paragraph{Finding IV: Reasoning in language aids models in logically linking visual cues.} From Figure~\ref{fig:gc-analysis}, it is evident that both CoT-normal and CoT-special, which reasoning in language, positively impact model performance in most cases. As demonstrated in Figure~\ref{fig:cot-special increase}, CoT-special improves performance by first having the model explicitly write out the cues present in each image, followed by using language to make inferences. This process helps reduce the model's error rate by structuring the task and providing clearer logical guidance. This suggests that when models are linking general visual cues, using language to help structure the logical flow of the process can be beneficial.

\paragraph{Finding V: Effectiveness of visual prompting depends on models' ability to interpret both prompting cues and the visual content.} 
As shown in Figure~\ref{fig:gc-analysis}, VP-grid negatively impacts GC performance for QwenVL2.5, causing a significant drop compared to the vanilla approach. Figure~\ref{fig:vp-grid decrease} reveals that this decline stems from the model's difficulty in interpreting the visual coordinates within the prompt, leading to misinterpretation of the cues and causing it to fail cases it originally answered correctly under the vanilla setting. However, as shown in Figure~\ref{fig:vp-grid increase}, GPT-4o successfully resolves a previously incorrect case by effectively leveraging the cues introduced through visual prompting while utilizing its strong visual perception abilities.

\subsection{Probing for Object-centric Cue (OC)}
\label{probing oc}
\paragraph{Methods.} (i) \textbf{CoT} (Table~\ref{cot prompt}), and (ii) \textbf{VP-zoom-o} (Figure~\ref{fig:VP-zoom-o}), which uses an open-set detector~\citep{grounded-sam} to obtain bounding boxes. These boxes are then cropped to focus the model’s attention on object-centric cues. By eliminating irrelevant non-object cues and emphasizing the object-centric cues, this approach enhances the model's ability to better focus on the most relevant visual information.

\paragraph{Finding VI: The open-ended nature of language may hinder object grouping.}
Unlike GC that link instance-level cues, OC requires grouping similar objects based on fine-grained visual details. As shown in Figure~\ref{fig:oc-analysis}, InternVL2.5 using CoT struggles with this task because the open-ended nature of language leads to both limited coverage of subtle visual cues (see Figure~\ref{fig:oc_cot_normal_decrease}) and inconsistent representations of the same cues, introducing ambiguity, making it harder for models to reliably align and group matching objects.

\paragraph{Finding VII: Amplifying object cues benefits stronger models while having minimal impact on others.} From Figure~\ref{fig:oc-analysis}, we observe that for models with strong vision capabilities like GPT-4o, our VP-zoom-o method further enhances performance. For other models, this method at least ensures that the performance remains on par with the vanilla approach, without causing any degradation.

\subsection{Probing for Person-centric Cue (PC)} 
\label{probing pc}
\paragraph{Methods.} (i) \textbf{CoT} (Table~\ref{cot prompt}). (ii) \textbf{VP-zoom-p} (Figure~\ref{fig:vp-zoom-p}) utilizes a face detector~\citep{facedetector} to obtain bounding boxes of faces-the most distinguishing feature of different individuals. It then crops the image to focus only on the face, thereby minimizing the interference from distractor cues such as clothing and other background elements.

\paragraph{Finding VIII: CoT and visual prompting fail to improve linking on highly abstract person-centric cues, leading to a performance drop.} From Figure~\ref{fig:pc-analysis}, we observe that for almost all models, neither CoT (language-based) nor VP-zoom-p (vision-based) lead to improved performance. This is because facial features are highly abstract, and CoT methods struggle to effectively describe them in words. Additionally, VP-zoom-p fails because current models' visual capabilities are insufficient to accurately perceive facial features.


\section{Related Work}

\paragraph{Advancements in vision-language models} have significantly broadened their capabilities ~\citep{gpt4o,Qwen2.5-VL,longva,llava-onevision,mplug-owl3,internvl,liang2024foundations}. Previously restricted to processing single-image inputs, many VLMs can now handle multi-image and even video inputs, allowing them to capture richer and more dynamic visual contexts. Additionally, with access to a growing volume of high-quality visual-textual paired training data~\citep{image-hq-1,image-hq2,image-hq-3,video-hq1,video-hq2,he2025mmboundaryadvancingmllmknowledge}, these models have shown substantial improvements in perceiving subtle visual cues and their relationships, enabling them to engage in more nuanced reasoning about visual content. Furthermore, VLMs are increasingly applied in real-world scenarios~\citep{navigation, motion, driving, ye2025survey, pi2024mllm}, solidifying their role in bridging vision and language for practical applications. However, to truly integrate into everyday life, VLMs still have significant room for improvement when it comes to more fundamental but common visual tasks, such as those assessed in our benchmark.

\paragraph{Benchmarking vision-language models} plays a critical role in guiding their future development~\citep{liang2024hemm,mme,mmstar}. These benchmarks typically focus on assessing the models' fine-grained perception~\citep{li2024naturalbench,mmvp}, reasoning abilities~\citep{scienceqa,mmvet,10787102}, commonsense knowledge~\citep{yue2024mmmu, wu-etal-2024-macaroon}, social intelligence~\citep{li2025mimeqasociallyintelligentnonverbalfoundation}, and robustness to input variations~\citep{fan2025v2rbenchholisticallyevaluatinglvlm}. In addition, evaluations targeting multi-image and video inputs are designed to measure the new competencies that VLMs require as their visual context extends. These tasks include captioning~\citep{yue2024mmmu,yu2019activitynet}, retrieval~\citep{wang2024muirbench,li2025migician}, comparison~\citep{MICBench,jiao2024img}, and temporal reasoning~\citep{liu2024tempcompass}.
However, existing benchmarks focus on evaluating VLMs' ability to interpret visual cues based on their knowledge. In contrast, humans typically solve such tasks by explicitly matching visual cues without relying on extensive background knowledge. To better assess whether they can replicate this human-like ability, we propose VLM2-Bench, which focuses on linking and matching explicit visual cues.
\section{Takeaways}

Based on our findings, we highlight three key areas for future improvements:

\begin{itemize}[leftmargin=*,noitemsep,topsep=0pt]
\item \textbf{Strengthening Fundamental Visual Capabilities.} Improving core visual abilities not only enhances overall performance but also increases adaptability. A stronger visual foundation maximizes the effectiveness of visual prompting and reduces reliance on prior knowledge, enabling models to operate more independently in vision-centric tasks.

\item \textbf{Balancing Language-Based Reasoning in Vision-Centric Tasks.} Integrating language into vision-centric tasks requires careful calibration. Future research should establish clearer principles on when language-based reasoning aids visual understanding and when it introduces unnecessary biases, ensuring models leverage language appropriately.

\item \textbf{Evolving Vision-Text Training Paradigms.} Current training paradigms focus heavily on emphasizing vision-language associations. However, as models expand their visual context window, their ability to reason purely within the visual domain becomes increasingly crucial. We should prioritize developing models that can structure, organize, and infer relationships among visual cues.
\end{itemize}
\section{Conclusion}
In summary, we introduce VLM2-Bench, a novel benchmark designed to probe the capability of vision-language models (VLMs) in visually linking matching cues, an essential yet underexplored skill for models in everyday visual reasoning. 
Through extensive evaluations and further analysis of prompting techniques applied on our benchmark, we identify 8 key findings. Based on these insights, we advocate for advancements in fundamental visual capabilities, better integration of language-based reasoning, and the evolution of vision-text training paradigms to improve VLMs' performance in vision-centric tasks.
\section*{Limitations}
VLM2-Bench focuses on evaluating visual cue linking but does not cover all possible scenarios. Additionally, while it provides valuable insights, its scale is limited, and model performance may not fully generalize to all real-world settings. Automated evaluation constraints limit the inclusion of open-ended questions in our benchmark, impacting the assessment of models' vision-centric reasoning abilities. Expanding task diversity and refining evaluation methods (e.g., switching the one-shot evaluation scenario to multi-turn conversations \citep{li2025personalized}) remain important directions for future work.
In future research, model self-play \cite{li2025mentalarenaselfplaytraininglanguage}, self-correction \cite{he2024selfcorrectionrefinementlearningframework}, and synthetic data pretraining \cite{qin2025scalinglawssyntheticdata} may also be interesting to explore.
\bibliography{custom}

\appendix
\section{Appendix Outline}
In the appendix, we provide:
\begin{itemize} \item \textbf{Appendix \ref{appendix: licence}} provides details on the licensing terms and usage rights for our benchmark.

\item \textbf{Appendix \ref{appendix: statistics}} presents the statistical analysis of the VLM2-Bench.

\item \textbf{Appendix \ref{appendix: baselines}} details on how we obtain the chance-level and human-level baselines.

\item \textbf{Appendix \ref{appendix: more details on benchmark construction}}  elaborates more details on the construction of the VLM2-Bench.

\item \textbf{Appendix \ref{appendix: prompting approaches}} provides a deeper dive into the various prompting techniques we use.

\item \textbf{Appendix \ref{Appendix: case study}} a detailed breakdown and analysis of failure and success examples regarding different prompting methods.
\end{itemize}

\section{Licencing and Intended Use}
\label{appendix: licence}
Our VLM2-Bench is available under the CC-BY 4.0 license for academic use with proper attribution. The images, videos, and annotations in this benchmark are intended solely for research purposes. These data were sourced from publicly available online platforms, and while efforts were made to use them responsibly, explicit permissions may not have been obtained for all content. Users are responsible for ensuring that their use of the data complies with applicable intellectual property laws and ethical guidelines. We encourage users to verify the sources and ensure compliance with any terms of service or licensing agreements.

\section{VLM2-Bench Statistics}
\label{appendix: statistics}



Here we provide additional details regarding the construction and statistics of our \textbf{VLM2-Bench} benchmark. As described in the main paper (\S~\ref{bench_statistics_main}), our benchmark comprises three main categories---\textit{General Cue (GC)}, \textit{Object-centric Cue (OC)}, and \textit{Person-centric Cue (PC)}---with a total of 3,060 visual-text query pairs. Below, we elaborate on the specific data composition, including the distribution of question types (T/F, multiple-choice (MC), numerical (Nu), and open-ended (Oe)) and the rationale behind each subtask.

\subsection{Overall Composition}
\begin{table}
\begin{tabular}{@{}lccccc@{}}
\toprule
\textbf{Category} & \textbf{T/F} & \textbf{MC} & \textbf{Nu} & \textbf{Oe} & \textbf{Total} \\
\midrule
GC & 960 & -- & -- & -- & 960 \\
OC & 720 & 200 & 360 & -- & 1,280 \\
PC & 400 & 100 & 120 & 200 & 820 \\
\midrule
Total & 2,080 & 300 & 480 & 200 & 3,060 \\
\bottomrule
\end{tabular}
\caption{Overview of query distribution across the three categories of VLM2-Bench. T/F = True/False, MC = multiple-choice, Nu = numerical, Oe = open-ended.}
\label{tab:overall_stats}
\end{table}
Table~\ref{tab:overall_stats} provides a detailed summary of the total query counts across different categories and subtasks in our benchmark. The dataset is structured into three primary categories: General Cue (GC), Object-centric Cue (OC), and Person-centric Cue (PC), comprising a total of 3,060 visual-text query pairs.

The General Cue (GC) category consists of 960 queries, which include 260 Matching (Mat) true/false pairs, resulting in 520 queries, and 220 Tracking (Trk) true/false pairs, leading to 440 queries. 

The Object-centric Cue (OC) category contains 1,280 queries, covering three subtasks: Comparison (Cpr) with 360 true/false pairs (720 queries), Counting (Cnt) with 360 numerical queries, and Grouping (Grp) with 200 multiple-choice questions. 

Lastly, the Person-centric Cue (PC) category includes 820 queries, comprising 200 Comparison (Cpr) true/false pairs (400 queries), 120 Counting (Cnt) numerical queries, 100 Grouping (Grp) multiple-choice questions, and 200 Free-form (VID) open-ended queries.

Overall, these components collectively sum up to 3,060 visual-text query pairs, offering a comprehensive benchmark for evaluating vision-language models across various types of contextual cues.

\subsection{Details per Subtask and Question Type}

\paragraph{General Cue (GC).}
\mbox{}\par
\noindent
\texttt{Matching (Mat).} We collect 260 True/False (T/F) pairs focused on verifying the alignment between a visual instance and a textual description (e.g., object presence, basic attributes). Each T/F pair forms two distinct queries (one True, one False), yielding 520 queries in total. 

\noindent
\texttt{Tracking (Trk).} We design 220 T/F pairs that test an understanding of object or entity continuity across frames. For example, a question might ask whether the same object reappears in subsequent frames. Each T/F pair similarly results in two queries, totaling 440.

\paragraph{Object-centric Cue (OC).} All the visual query cases are built upon the 360 image sequences we construct. Details about image sequences can be found in Section~\ref{appendix_oc_construct}.

\noindent
\texttt{Comparison (Cpr).} This subtask examines the model's ability to compare object properties (e.g., size, color, quantity) across different frames. We produce 360 T/F pairs, each yielding two queries (720 total). Among these 360 pairs, we maintain a 1:2 ratio of True to False for ground-truth answers (i.e., 120 True vs.\ 240 False).

\noindent
\texttt{Counting (Cnt).} We provide 360 numerical questions, each asking for a count of objects in a given scene or sequence. Possible numeric answers are typically small integers (e.g., 1, 2, 3), reflecting the number of relevant objects.

\noindent
\texttt{Grouping (Grp).} We generate 200 multiple-choice (MC) questions that ask about grouping objects according to certain criteria (e.g., AAB, ABC, AAAB, AABC, ABCD). Each question presents multiple group-configuration options plus a \textit{``None''} option, which can serve as either a correct or distractor choice. For image sequences of length 4, the options include various plausible groupings (two-of-a-kind, three-of-a-kind, etc.) along with at least one additional distractor grouping that also involves three-of-a-kind to ensure sufficient challenge.

\paragraph{Person-centric Cue (PC).} Similar to OC, the construction of 260 image sequences as well as 200 video clips for PC is detailed in Section~\ref{appendix_pc_construct}.

\par

\noindent
\texttt{Comparison (Cpr).} We create 200 T/F pairs (400 queries total) focusing on comparing attributes or actions related to one or more human individuals across multiple images in a sequence. The ground truth is balanced at 100 True vs.\ 100 False.

\noindent
\texttt{Counting (Cnt).} This subtask involves 120 numerical questions asking for the number of people present or the frequency of certain actions in a sequence. Typical numeric answers range from 1 to 4, given the scope of each visual sequence.

\noindent
\texttt{Grouping (Grp).} We provide 100 MC questions based on sequences containing at least three images, with at least two images featuring the same main ``meta-human.'' The goal is to identify correct groupings of persons based on appearance, role, or action. As with \textit{OC-Grp}, each question includes a ``None'' option as either the correct or a distractor choice.

\noindent
\texttt{Open-ended (VID).} We introduce 200 open-ended queries that focus on various person-centric aspects, such as identifying roles or describing activities. These questions allow more flexibility in model responses and assess the ability to generate context-relevant answers.

\subsection{Annotation Quality and Agreement}
As noted in the main text, three annotators reviewed all 3,060 question-answer pairs. An inter-annotator agreement study showed a high consensus rate of 98.74\%, ensuring that the data is both accurate and consistent.

\subsection{Summary}
Our construction methodology ensures a balanced coverage of both object-centric and person-centric reasoning, as well as basic general cues such as element matching and tracking. The inclusion of multiple question types (T/F, MC, numerical, and open-ended) further promotes comprehensive evaluation of vision-language models. Figure~\ref{fig:bench statistics main} in the main paper illustrates the distribution of these subtasks and their question-format breakdown. We believe that the richness and diversity of VLM2-Bench make it a robust platform for advancing multimodal research.

\section{Baselines}
\label{appendix: baselines}
\subsection{Chance-level}
In this part, we explain the calculation of chance-level accuracy for all subtasks in Table~\ref{exp:main_exp}.

\paragraph{GC-Mat, GC-Trk.} 
The Matching (Mat) and Tracking (Trk) tasks in General Cue (GC) follow a \textbf{True-False (TF) paired-question format}, where each pair consists of a \textbf{positive question} and a \textbf{negative question}:

\begin{itemize}
    \item \textbf{Positive Question}: Derive from the correct \textit{element} or \textit{change}.
    The ground truth (GT) answer is True (T).
    \item \textbf{Negative Question}: Derive from the distractor \textit{element} or \textit{change}.
    The ground truth (GT) answer is False (F).
\end{itemize}

A question pair example is shown in Table~\ref{tf_pair_example}.

\begin{table}[h]
    \begin{tcolorbox}[colframe=black, colback=gray!10!white, coltitle=black, boxrule=0.5mm]
    \textbf{Positive Question:}  
    
    \textit{"Is the answer \textcolor{customgreen}{`the salad'} correct for the given question: 'What object that was present in the first image is no longer visible in the second?'"}  
    \\ GT Answer: \textbf{\textcolor{customgreen}{T}}

    \vspace{3mm}
    
    \textbf{Negative Question:}  
    
    \textit{"Is the answer \textcolor{customred}{`the ciabatta roll'} correct for the given question: 'What object that was present in the first image is no longer visible in the second?'"}  
    \\ GT Answer: \textbf{\textcolor{customred}{F}}
    \end{tcolorbox}
    \caption{Example of True-False paired questions in GC-Mat, with a positive and negative question.}
    \label{tf_pair_example}
\end{table}

During the construction of these questions, we ensure that the queried content originates from either the correct answer or a distractor answer. These elements are designed to be \textbf{independent and identically distributed}. Since each question in the pair has an independent 50\% chance of being answered correctly, the expected accuracy under random guessing would be $P(\text{correct answer}) = \frac{1}{2} \times \frac{1}{2} = \frac{1}{4} = 25\%$.

\paragraph{OC-Cpr, PC-Cpr.} 

The OC-Cpr and PC-Cpr tasks utilize a \textbf{True-False (TF) paired-question format} where both questions in a pair originate from the same correct answer but are framed in two different ways:

\begin{itemize}
    \item \textbf{Positive Question}: A direct affirmative statement that correctly represents the ground truth.
    \item \textbf{Negative Question}: A negated version of the positive question, often by inserting "not" after the verb.
\end{itemize}

An example is shown in Table~\ref{tf_cpr_example}.

\begin{table}[h]
    \begin{tcolorbox}[colframe=black, colback=gray!10!white, coltitle=black, boxrule=0.5mm]
    \textbf{Positive Question:}  
    
    \textit{"Given the images, the claim `The pets in these images \textcolor{customgreen}{are} the same pet.' is right."}  
    \\ GT Answer: \textbf{\textcolor{customgreen}{T}}

    \vspace{3mm}

    \textbf{Negative Question:}  
    
    \textit{"Given the images, the claim `The pets in these images \textcolor{customred}{are not} the same pet.' is right."}  
    \\ GT Answer: \textbf{\textcolor{customred}{F}}
    \end{tcolorbox}
    \caption{Example of True-False paired questions in OC-Cpr, with a positive and negative question.}
    \label{tf_cpr_example}
\end{table}

This construction aims to eliminate \textbf{language bias} by ensuring that the model does not favor one phrasing over another. For a language model that is free from bias, these two questions are \textbf{logically equivalent}—answering one correctly implies answering the other correctly as well. Consequently, under random guessing, the expectation is $P(\text{correct answer}) = \frac{1}{2} = 50\%$.

\paragraph{OC-Cnt, PC-Cnt.}
The calculation formulas for the accuracy of the chance-level accuracy are the same as in Section \ref{metrics}.

Under a pure random guessing strategy, the predicted answer \( \hat{N}_i \) is uniformly sampled from the set \(\{1,2,\ldots,L\}\), where \(L\) is the number of images (i.e., the sequence length for that instance). For a fixed sequence length \(L\), we can compute the expected normalized accuracy \(E(L)\) by averaging over all possible ground-truth and guess pairs:
\[
E(L) = 1 - \frac{1}{L^2} \sum_{N=1}^{L} \sum_{\hat{N}=1}^{L} w(L) \cdot \epsilon(N,\hat{N})^{\alpha},
\]
where
\[
\epsilon(N,\hat{N}) = \frac{|\hat{N} - N|}{\max(N-1,\, L-N)}
\]
and the weight is defined as
\[
w(L) = \frac{L_{\max}}{L},
\]
with \( L_{\max} = 4 \) being the maximum sequence length in our dataset.

\textbf{OC-Cnt Task:} The OC-Cnt task exhibits the following distribution:
\begin{itemize}
    \item Length 2: 80 sequences (22.2\%)
    \item Length 3: 120 sequences (33.3\%)
    \item Length 4: 160 sequences (44.4\%)
\end{itemize}
Thus, the overall chance level accuracy is obtained as the weighted average:
$
Acc_{\text{OC-Cnt}} = \frac{80\,E(2) + 120\,E(3) + 160\,E(4)}{360} \approx 34.88\%.
$

\textbf{PC-Cnt Task:} For the PC-Cnt task, the sequence distribution is:
\begin{itemize}
    \item Length 2: 30 sequences (25.0\%)
    \item Length 3: 25 sequences (20.8\%)
    \item Length 4: 65 sequences (54.2\%)
\end{itemize}
Accordingly, the overall chance level accuracy is given by:
$
Acc_{\text{PC-Cnt}} = \frac{30\,E(2) + 25\,E(3) + 65\,E(4)}{120} \approx 34.87\%.
$


\subsection{Human-level}
\begin{figure*}[htbp]
  \centering
  \includegraphics[width=0.85\textwidth]{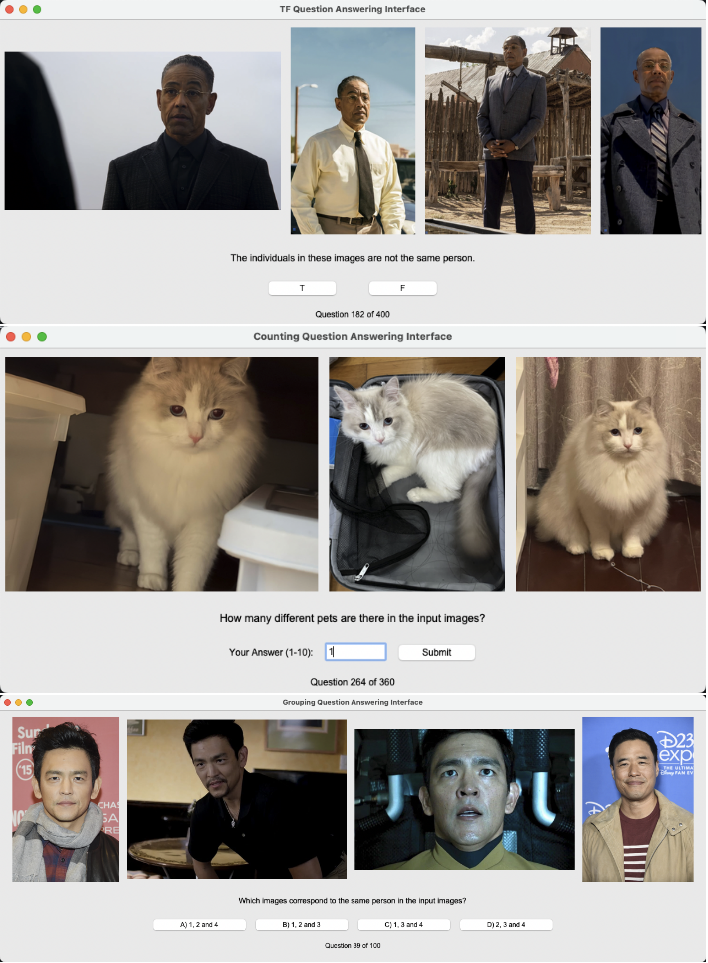} 
  \caption{The GUI used for human-level testing.}
  \label{fig: gui_human}
\end{figure*}

To facilitate human participants in providing responses to our questions, we integrated all model-prompted questions and answer choices into a graphical user interface (GUI), as illustrated in Figure~\ref{fig: gui_human}. This interface enabled participants to select their answers conveniently, ensuring consistency in data collection. We then gathered all responses and conducted statistical analysis on the collected human evaluations.

\section{More details on Benchmark Construction}
\label{appendix: more details on benchmark construction}

\subsection{GC (General Cue)}
\paragraph{Manual Screening and Refine.}  

\begin{figure*}[htbp]
  \centering
  \includegraphics[width=0.85\textwidth]{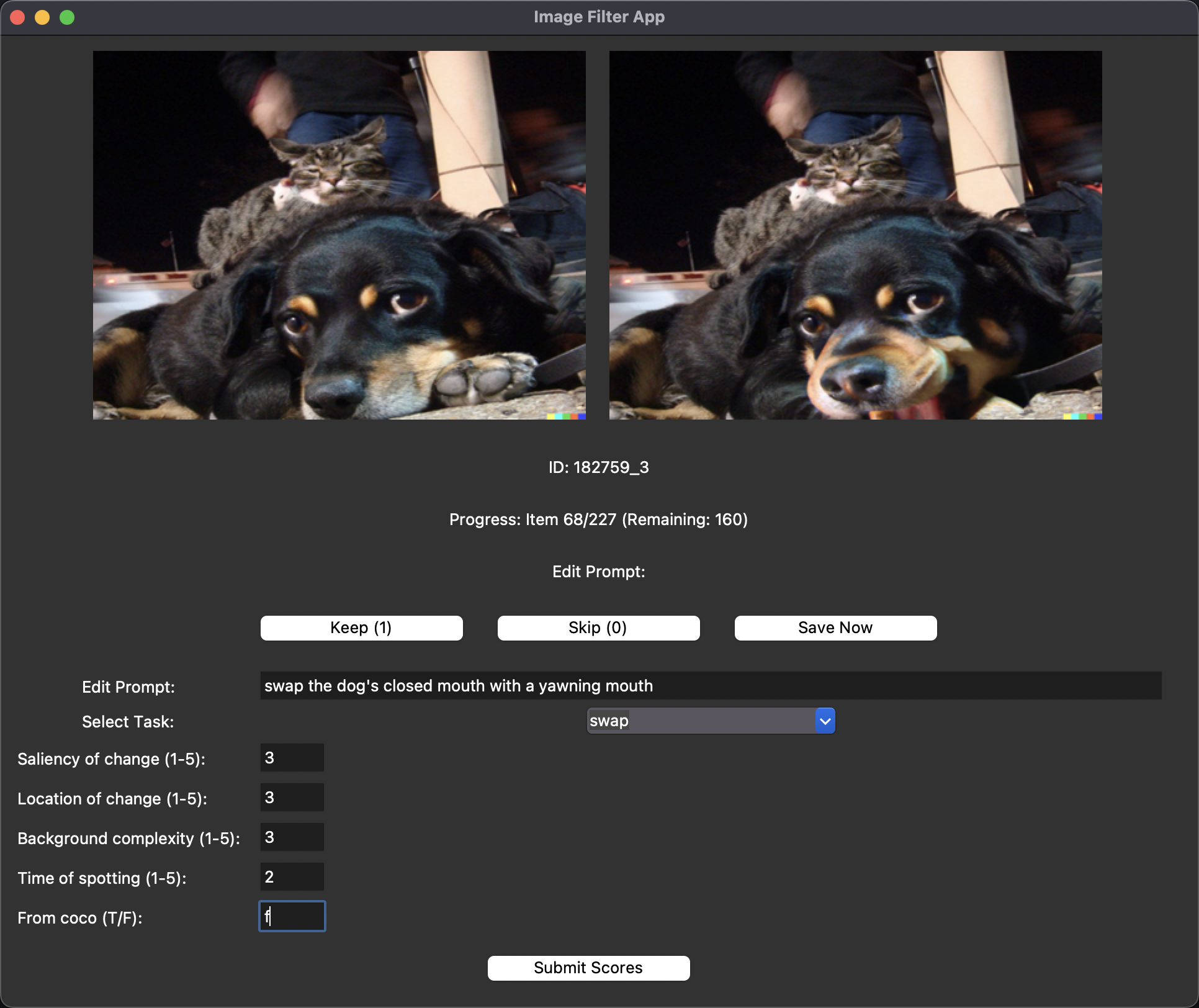} 
  \caption{The GUI used for manually screening image editing data and refining edited prompts in General Cue (GC).}
  \label{fig: gui}
\end{figure*}

Figure~\ref{fig: gui} demonstrates the Graphic User Interface (GUI) we build for manually screening image editing data.

\paragraph{Salient Sampling.}

The pseudocode in Figure~\ref{fig:salient score algorithm} and Table~\ref{template salient score} displays the calculation process for the salient sampling score mentioned in Section~\ref{gc}.

\begin{table*}[htbp]
    \begin{tcolorbox}[colframe=black, colback=gray!10!white, coltitle=black, boxrule=0.5mm]
    Supposed you are looking at two images:

    Image 1: \textcolor{teal!70}{\textbf{<Cap\_{src}>}}

    Image 2: \textcolor{orange!70}{\textbf{<Cap\_{edit}>}}

    From Image 1 to Image 2, the change can be summarized as: \textcolor{blue!70}{\textbf{<P>}}
    \end{tcolorbox}
    \caption{Template for salient-score calculation, which contain three placeholders for each sample.}
    \label{template salient score}
\end{table*}

\begin{figure*}[htbp]
\centering
\begin{minipage}{0.90\textwidth} 
    \begin{algorithm}[H]
    \caption{\footnotesize Salient Score Computation}
    \begin{lstlisting}[language=Python]
# cap_src: caption for the source image
# cap_edit: caption for the edited image
# T: template for constructing a paragraph
# P: editing prompt
input_text = concat(cap_src, cap_edit, T)
in_tokens = tokenizer.encode(input_text)
out_tokens = tokenizer.encode(P)
log_sum = 0
tokens = in_tokens

# Model Forward Pass
for i in range(1, len(out_tokens)):
    outputs = model(tokens)
    logits = outputs.logits

    # Extract log probability of next token
    probs = log_softmax(logits[0, -1, :])
    prob = probs[out_tokens[i]]
    log_sum += prob
    
    # Update Input Sequence
    tokens = concat(tokens, out_tokens[i])

# Normalize the total log probability as the salient_score
salient_score = log_sum / len(out_tokens)

# Return: salient_score
    \end{lstlisting}
\end{algorithm}
\end{minipage}
\caption{Pseudocode for salient score computation in the phrase of Salient Sampling in the construction of GC.}
\label{fig:salient score algorithm}
\end{figure*}

\paragraph{Prompts for Pair-wise Answer Generation.}

Table~\ref{tab:mat_pair_generation_prompt} and \ref{tab:trk_pair_generation_prompt}
provides the complete prompts used to generate pair-wise answers for our evaluation tasks. The prompts were designed to instruct the language model to produce two distinct answers—a positive (T) answer and a negative (F) answer—for each task. The dual-answer format is intended to capture both the expected response and its direct opposite, thereby offering a more balanced insight into the model's understanding.

\begin{table*}[htbp]
    \begin{tcolorbox}[colframe=black, colback=gray!10!white, coltitle=black, boxrule=0.5mm]
\textbf{\#Task Description}

Given the change between the first image and the second image, you need to generate four choices to the question ``What new element can be observed in the second image that was not present in the first?" \textcolor{blue}{(this question varies based on the mis-matched cue types, here shows the question for `add" from the ``Add/Remove" category)}. Remember, the choices' lengths should be similar. Additionally, your response should start with "\textit{Choices:}".

\textbf{\#Pair Design}

In these two choices, you need to contain *only* the names of objects, but be specific:

1. Correct Answer (You need to infer the *only* from the \textit{Editing Information})

2. Distractor (You need to pick a random object *only* in the \textit{Description}, but differ from the correct answer object)

\textbf{\#In-context example}

\textit{Editing Information}: 

Add a katana held in the figure's left hand, angled downwards.

\textit{Description}: 

The image depicts a person dressed in traditional Japanese armor, standing in a misty, snowy landscape. The armor is detailed and appears to be made of metal, with various straps and buckles. The person is wearing a black mask that covers their entire face, adding to the mysterious and stealthy appearance. The background features stone lanterns and other traditional Japanese structures, which are partially obscured by the mist. The overall atmosphere is serene yet somewhat eerie, with the mist adding a sense of mystery and isolation. The scene suggests a historical or fantasy setting, possibly a samurai or ninja in a snowy, misty environment.

\textit{Choices}:

Correct Answer: katana held

Distractor: black mask

\textbf{\#Task}

\textit{Editing Information}: 

\textcolor{teal}{\textbf{<Edit Prompt>}}

\textit{Description}: 

\textcolor{orange}{\textbf{<Description>}}

    \end{tcolorbox}
    \caption{Prompt for generating paired answers in the Matching (Mat) subtask of General Cue (GC).}
    \label{tab:mat_pair_generation_prompt}
\end{table*}

\begin{table*}[htbp]
    \begin{tcolorbox}[colframe=black, colback=gray!10!white, coltitle=black, boxrule=0.5mm]
\textbf{Task Description}

Given the change between the first image and the second image, you need to generate four choices to the question "What key visual difference can be observed from the first image to the second image?". Remember, the choices' lengths should be similar. Additionally, your response should start with ``\textit{Choices}:" and must contain Correct Answer and Direct Reverse Answer.

\textbf{Pair Design}

In the two choices, you need to contain:

1. Correct Answer (You need to infer from the \textit{Editing Information})

2. Direct Reverse Answer (You need to infer from the \textit{Editing Information} and change it to the opposite)

\textbf{In-context example} 

\textit{Editing Information}: 

Swap the black ninja gloves with clean white gloves appropriate for serving.

\textit{Description}: 

The image depicts a person dressed in formal attire, standing in a doorway. The individual is wearing a black tuxedo with a white dress shirt and a black bow tie. They are holding a tray with several items on it. The tray contains a small glass container, a bottle, and a small white object, possibly a salt shaker or a similar item. The person is also wearing black gloves, which are typical for serving or formal dining scenarios. The background shows a wooden door with a brass hinge and a light-colored wall. The setting appears to be indoors, possibly in a house or a formal establishment.

\textit{Choices}:

Correct Answer: The black ninja gloves were replaced with clean white gloves.

Direct Reverse Answer: The clean white gloves were replaced with black ninja gloves.

\textbf{\#Task}

\textit{Editing Information}: 

\textcolor{teal}{\textbf{<Edit Prompt>}}

\textit{Description}: 

\textcolor{orange}{\textbf{<Description>}}

    \end{tcolorbox}
    \caption{Prompt for generating paired answers in the Tracking (Trk) subtask of General Cue (GC).}
    \label{tab:trk_pair_generation_prompt}
\end{table*}

\subsection{OC (Object-centric Cue)}
\label{appendix_oc_construct}

\begin{figure*}[htbp]
  \centering
  \includegraphics[width=0.99\textwidth]{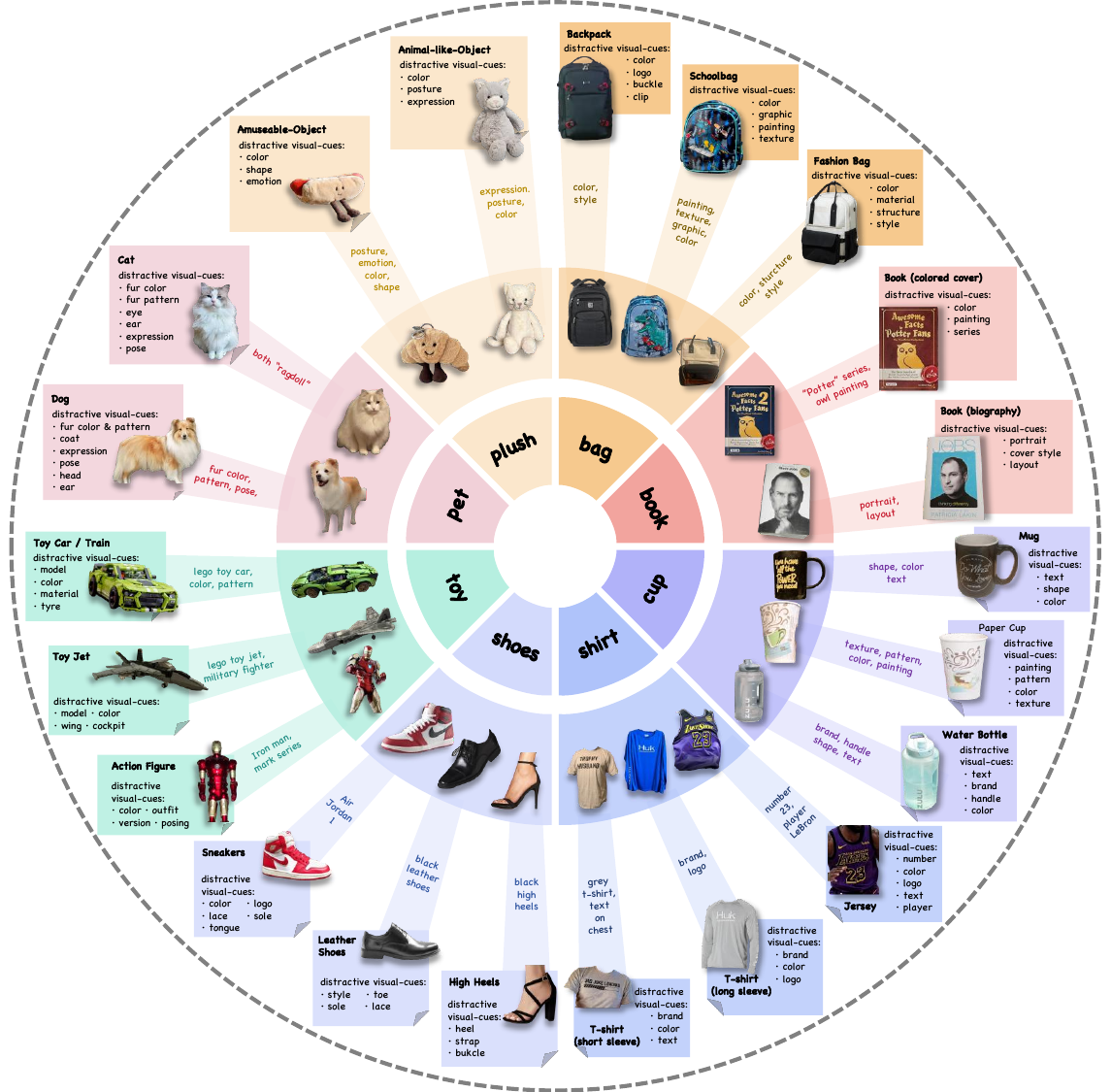} 
  \caption{The overview of the structured design of the Object-centric Cue (OC) images.
 \textbf{Central Layer (Main Meta-Objects)}: The innermost circle represents the predefined \textbf{8 object categories}, which serve as the foundation for our dataset. These categories include \textit{Pet, Plush, Bag, Book, Cup, Shirt, Shoes, and Toy}. Each category consists of 4 main meta-objects.
    \textbf{Middle Layer (Example Meta-Objects within Each Category)}: Each segment surrounding the center showcases a representative \textbf{main meta-object} within its category. These meta-objects serve as core instances for data collection. For example, the \textit{Pet} category includes \textit{Cat} and \textit{Dog}, while the \textit{Bag} category includes \textit{Backpack}, \textit{Schoolbag} and \textit{Fashion Bag}.
    \textbf{Outer Layer (Distractor Meta-Objects \& Visual Cue Distraction Principles)}: The outermost ring presents 1 out of 4 \textbf{distractor meta-objects} specifically selected to create challenging image sequences. Each distractor meta-object shares one or more \textbf{distractive visual cues} with its corresponding main meta-object.
  }
  \label{fig:oc overview}
\end{figure*}

\paragraph{Data Collection.} 
To construct the dataset, we follow a structured approach to collect object-centric images, as illustrated in Figure~\ref{fig:oc overview}. In total, we manually collected 320 images for objects.

\paragraph{Main Meta-Object Selection.} 
We predefine 8 types of common objects, with each type containing 5 meta-objects to ensure a class-balanced sampling and avoid long-tail distribution \citep{Yao_2023_ICCV}. For each meta-object, we collect four images that represent the same object from different angles and scene conditions.

\paragraph{Distractor Meta-Object Selection.} 
To build meaningful object image sequences, we introduce visually distractive elements for each main meta-object, referred to as ``distractor meta-objects''. Specifically, for each main meta-object, we collect four additional images that belong to different but visually similar meta-objects within the same object category. These images are selected following predefined visual cue confusion principles, ensuring that they provide meaningful challenges for vision language models. We ensure that each distractor image belongs to a different distractor meta-object, fundamentally guaranteeing that the count of different meta-objects in the final constructed sequence strictly follows our design. The principle of selecting distractor meta-objects is illustrated in the outer ring of Figure~\ref{fig:oc overview}.

\paragraph{Image Sources.} 
The images are gathered from various sources based on the nature of the objects:
\begin{itemize}
    \item \textbf{Plush Objects:} Images of plush toys are entirely sourced from the \href{https://us.jellycat.com/}{Jellycat website} and its review sections, where diverse user-uploaded images provide a wide variety of object angles and scenes.
    \item \textbf{Pet Objects:} For the pet category of meta-objects, we source images from a combination of social media accounts of popular pet influencers' pet photography. We also include images of a ragdoll cat owned by one of the authors. As a result, this approach guarantees that each pet meta-object within the dataset belongs to the same individual cat or dog, minimizing variability unrelated to visual cue confusion.
    \item \textbf{Other Objects:} Most images are collected from \href{https://www.amazon.com/}{Amazon} product listings and review sections containing user-uploaded photos. A smaller portion of the dataset is curated using Google Lens image search, where specific visual distractive cues are used to retrieve and manually select images. The detailed visual cue principles guiding this selection process can be found in Figure~\ref{fig:oc overview}.
\end{itemize}

\paragraph{Images Sequence Construction.} 
The construction of image sequences in OC (a total of 360 sequences) follows the structure in Table~\ref{tab:sequence construction oc}. More specific details are listed below:

\textbf{Two-Image Sequences (\texttt{image\_seq\_len} = 2)}
\begin{enumerate}
    \item \textbf{Main Meta-Object Only (AA)}:
        Two images are randomly sampled from the same main meta-object.  
        40 sequences are constructed (one for each main meta-object).
    \item \textbf{Main Meta-Object + Distractor Meta-Object (AB)}:
        One image is randomly selected from the main meta-object, and one from the corresponding distractor meta-object.  
        40 sequences are constructed.
\end{enumerate}

\textbf{Three-Image Sequences (\texttt{image\_seq\_len} = 3)}
\begin{enumerate}
    \item \textbf{Main Meta-Object Only (AAA)}:
        Three images are randomly sampled from the same main meta-object.  
        40 sequences are constructed.
    \item \textbf{Main Meta-Object + Distractor Meta-Object (AAB)}:
        Two images are selected from the main meta-object, and one from the distractor meta-object.  
        The order of images is shuffled.  
        40 sequences are constructed.
    \item \textbf{Main Meta-Object + Distractor Meta-Objects (ABC)}:
        One image is selected from the main meta-object, while two are selected from different distractor meta-objects.  
        40 sequences are constructed.
\end{enumerate}

\textbf{Four-Image Sequences (\texttt{image\_seq\_len} = 4)}
\begin{enumerate}
    \item \textbf{Main Meta-Object Only (AAAA)}:
        All four images are sampled from the same main meta-object and shuffled.  
        40 sequences are constructed.
    \item \textbf{Main Meta-Object + Distractor Meta-Object (AAAB)}:
        Three images are sampled from the same main meta-object, while one is selected from a distractor meta-object.  
        40 sequences are constructed.
    \item \textbf{Main Meta-Object + Distractor Meta-Objects (AABC)}:
        Two images are selected from the main meta-object, while two are selected from different distractor meta-objects.  
        40 sequences are constructed.
    \item \textbf{Main Meta-Object + Distractor Meta-Objects (ABCD)}:
        One image is selected from the main meta-object, while three are selected from different distractor meta-objects.  
        40 sequences are constructed.
\end{enumerate}

\begin{table*}[h]
\centering
\begin{tabularx}{\textwidth}{ccYccc}
\toprule
\textbf{Num} & \textbf{Src} & \textbf{Process of Image Sequences Construction} &  \textbf{\textit{Cpr}} & \textbf{\textit{cnt}} & \textbf{\textit{Grp}} \\ \hline
\multirow{2}{*}{2} & \multirow{2}{*}{\textbf{AA}} & 
2 images from the same object \( O_i \), randomly sampled as \( \mathcal{I}_{O_i} = \{ I_i, I_j \} \), and shuffled. 
& \multirow{2}{*}{T} & \multirow{2}{*}{2} & \multirow{2}{*}{-}  \\ \hline
\multirow{2}{*}{2} & \multirow{2}{*}{\textbf{AB}} & 
1 image \( I_{i} \) from \( \mathcal{I}_{O_i} \) and 1 image \( I_{\neg i} \) from distractor set \( \mathcal{I}_{\neg O_i}\), randomly shuffled. 
& \multirow{2}{*}{F} & \multirow{2}{*}{1} & \multirow{2}{*}{-}  \\ \hline
\multirow{2}{*}{3} & \multirow{2}{*}{\textbf{AAA}} & 
3 images from the same object \( O_i \), randomly sampled as \( \mathcal{I}_{O_i} = \{ I_i, I_j, I_k \} \), and shuffled. 
& \multirow{2}{*}{T} & \multirow{2}{*}{3} & \multirow{2}{*}{-}  \\ \hline
\multirow{3}{*}{3} & \multirow{3}{*}{\textbf{AAB}} & 
2 images from the same object \( O_i \), randomly sampled as \( \mathcal{I}_{O_i} = \{ I_i, I_j\} \) and 1 \( I_{\neg i} \) from distractor set \( \mathcal{I}_{\neg O_i}\), randomly shuffled. 
& \multirow{3}{*}{F} & \multirow{3}{*}{2} & \multirow{3}{*}{[$I_i$, $I_j$]}  \\ \hline
\multirow{3}{*}{3} & \multirow{3}{*}{\textbf{ABC}} & 
1 images from the same object \( O_i \), randomly sampled as \( \mathcal{I}_{O_i} = \{ I_i\} \) and 2 images \( \{I_{\neg i}, I_{\neg j}\} \) from distractor set \( \mathcal{I}_{\neg O_i}\), randomly shuffled. 
& \multirow{3}{*}{F} & \multirow{3}{*}{3} & \multirow{3}{*}{[]}  \\ \hline
\multirow{2}{*}{4} & \multirow{2}{*}{\textbf{AAAA}} & 
4 images from the same object \( O_i \), randomly sampled as \( \mathcal{I}_{O_i} = \{ I_i, I_j, I_k, I_p  \} \), and shuffled. 
& \multirow{2}{*}{T} & \multirow{2}{*}{4} & \multirow{2}{*}{-}  \\ \hline
\multirow{3}{*}{4} & \multirow{3}{*}{\textbf{AAAB}} & 
3 images from the same object \( O_i \), randomly sampled as \( \mathcal{I}_{O_i} = \{ I_i, I_j, I_k \} \) and 1 image \( I_{\neg i} \) from distractor set \( \mathcal{I}_{\neg O_i}\), randomly shuffled. 
& \multirow{3}{*}{F} & \multirow{3}{*}{2} & \multirow{3}{*}{[\(I_i, I_j, I_k\)]}  \\ \hline
\multirow{3}{*}{4} & \multirow{3}{*}{\textbf{AABC}} & 
2 images from the same object \( O_i \), randomly sampled as \( \mathcal{I}_{O_i} = \{ I_i, I_j\} \) and 2 images \( \{I_{\neg i}, I_{\neg j}\} \) from distractor set \( \mathcal{I}_{\neg O_i}\), randomly shuffled. 
& \multirow{3}{*}{F} & \multirow{3}{*}{3} & \multirow{3}{*}{[\(I_i, I_j\)]}  \\ \hline
\multirow{3}{*}{4} & \multirow{3}{*}{\textbf{ABCD}} & 
1 images from the same object \( O_i \), randomly sampled as \( I_i\) and 3 images \( \{I_{\neg i}, I_{\neg j}, I_{\neg k} \} \) from distractor set \( \mathcal{I}_{\neg O_i}\), randomly shuffled. 
& \multirow{3}{*}{F} & \multirow{3}{*}{3} & \multirow{3}{*}{[]}  \\ \bottomrule
\end{tabularx}
\caption{Summary of multi-images sequence construction for Object-centric Cue (OC) tasks.}
\label{tab:sequence construction oc}
\end{table*}

\paragraph{Question Templates.}

Table \ref{oc_cpr_tf_example}, \ref{oc_cnt_example} and \ref{oc_grp_example} list detailed standard question templates (with format instructions) for the Object-centric Cue task, including 3 subtasks: Comparison (cpr), Counting (Cnt), and Grouping (Grp). 

\begin{table}[htbp]
    \begin{tcolorbox}[colframe=black, colback=gray!10!white, coltitle=black, boxrule=0.5mm]
    \textbf{OC-Cpr Positive Question:}  
    
    \textit{Judge the following statement based on the images: `The \{obj\}s in these images are the same \{obj\}.' Provide only one correct answer: `T' (True) or `F' (False). Respond with either `T' or `F'.}  
    \\ GT Answer: \textbf{\textcolor{customgreen}{T}}

    \vspace{3mm}
    
    \textbf{OC-Cpr Negative Question:}  
    
    \textit{Judge the following statement based on the images: `The \{obj\}s in these images are \textcolor{customred}{not} the same \{obj\}.' Provide only one correct answer: `T' (True) or `F' (False). Respond with either `T' or `F'.}   
    \\ GT Answer: \textbf{\textcolor{customred}{F}}
    \end{tcolorbox}
    \caption{Question templates used for consistency-pair evaluation in the Comparison (Cpr) subtask of Object-centric Cue (OC).}
    \label{oc_cpr_tf_example}
\end{table}

\begin{table}[htbp]
    \begin{tcolorbox}[colframe=black, colback=gray!10!white, coltitle=black, boxrule=0.5mm]
    \textbf{OC-Cnt Question:}  

    \textit{Answer the following question according to this rule: You only need to provide *ONE* correct numerical answer. For example, if you think the answer is `1', your response should only be `1'. The Question is: How many different \{obj\}s are there in the input images?}  
    \\ GT Answer: \textbf{3} \textcolor{blue}{(Example Answer)}
    \end{tcolorbox}
    \caption{The question template used for the counting (Cnt) subtask of Object-centric Cue (OC).}
    \label{oc_cnt_example}
\end{table}

\begin{table}[htbp]
    \begin{tcolorbox}[colframe=black, colback=gray!10!white, coltitle=black, boxrule=0.5mm]
    \textbf{OC-Grp Question:}  

    \textit{Answer the following question based on this rule: You only need to provide *ONE* correct answer, selecting from the options listed below. For example, if you think the correct answer is `B) 1 and 2', your response should be `B) 1 and 2'. \\
    The Question is: Which images show the same \{obj\} in the input images? Choices: A) 1 and 3; B) None; C) 2 and 3; D) 1 and 2.}  
    \\ GT Answer: \textbf{A) 1 and 3} \textcolor{blue}{(Example Answer)}
    \end{tcolorbox}
    \caption{The question template used for the grouping (Grp) subtask of Object-centric Cue (OC).}
    \label{oc_grp_example}
\end{table}

\subsection{PC (Person-centric Cue)}

\label{appendix_pc_construct}

\paragraph{Data Collection.}
We collect images of \emph{meta-humans} mainly from \url{https://www.imdb.com/} and some are from the actor or actress's social media.

\paragraph{Main Meta-human Selection.} 

Our dataset is evenly distributed across different racial groups (Asian, Black, and White) and genders (Male and Female). 
For every race-gender combination, we select five main meta-humans, each contributing four images, yielding a total of 120 images. 

To ensure consistency, all selected individuals are within a similar age range, preventing significant age-related facial changes that could interfere with identity recognition. Additionally, each actor's appearance remains relatively consistent in terms of makeup and overall styling, ensuring that different images of the same meta-human retain distinct yet comparable visual cues (e.g. face shape, eye spacing, nose structure, and lip contours). By preserving these features, we avoid manipulating a single individual's visual cues that could potentially mislead VLMs. Rather, we ensure that the evaluation genuinely tests whether the model can visually link matching cues to recognize the same or different individuals without prior identity knowledge.

\paragraph{Distractor Meta-human Selection.}
To introduce challenging distractors in our sequences, we compute the CLIP embedding for every image and store these embeddings in a reference base. 
When a distractor image is needed, we perform an image-to-image similarity search within this base to identify the most visually similar image that originates from a different meta-human. 
This fine-grained matching ensures that the distractor image closely resembles the main meta-human’s image, leading to more challenging image sequences.

\paragraph{Discussion on Why Objects Require Dedicated Distractors, While Humans Do Not.}  
In object-centric tasks, objects are categorized into eight distinct types, with substantial differences among different types (e.g. pets and bags). Therefore, each main meta-object requires dedicated distractors from the same object type to ensure meaningful comparisons.  
In contrast, humans belong to a single category, meaning that any meta-human can serve as a distractor for another. Given that we compute CLIP embeddings to select visually similar distractors, the constructed image sequences already present a significant challenge without the need for type-specific distractors.  
We also ensure diversity by selecting five main meta-humans for each race-gender pair, providing a sufficiently large pool from which to choose suitable distractors. Corresponding to our hypothesis, in the final curated sequences, most distractor meta-humans chosen were of the same race or gender as the main meta-human. Additionally, as shown in Table \ref{exp:main_exp}, these curated image sequences along with our designed questions effectively challenge tested models, revealing their limited performances in visually linking matching cues on person-centric data.

\paragraph{Images Sequence Construction.}
The construction of image sequences in PC (a total of 260 sequences) follows the structure in Table~\ref{tab:sequence construction pc}. 
More specific details are listed below:

\textbf{Two-Image Sequences (\texttt{image\_seq\_len} = 2)}
\begin{enumerate} \item \textbf{Main Meta-Human Only (PP):}
Two images are randomly selected from the same main meta-human, resulting in 50 sequences.
\item \textbf{Main Meta-Human + Distractor Meta-Human (PQ):}  
One image is randomly selected from the main meta-human, and the other from a distractor meta-human. The order of the images is shuffled. This results in 50 sequences.

\end{enumerate}

\textbf{Three-Image Sequences (\texttt{image\_seq\_len} = 3)}
\begin{enumerate}
    \item \textbf{Main Meta-Human Only (PPP)}:  
    Three images are randomly sampled from the same main meta-human.  
    20 sequences are constructed.
    \item \textbf{Main Meta-Human + Distractor Meta-Human (PPQ)}:  
    Two images are selected from the main meta-human, and one from a single distractor meta-human.  
    The order of images is shuffled.  
    30 sequences are constructed.
    \item \textbf{Main Meta-Human + Distractor Meta-Humans (PQR)}:  
    One image is selected from the main meta-human, while the other two come from distinct distractor meta-humans.  
    The order is shuffled.  
    10 sequences are constructed.
\end{enumerate}

\textbf{Four-Image Sequences (\texttt{image\_seq\_len} = 4)}
\begin{enumerate}
    \item \textbf{Main Meta-Human Only (PPPP)}: 
    All four images are sampled from the same main meta-human.   
    30 sequences are constructed.
    \item \textbf{Main Meta-Human + Distractor Meta-Human (PPPQ)}:  
    Three images are sampled from the main meta-human, while one is selected from a single distractor meta-human.  
    20 sequences are constructed.
    \item \textbf{Main Meta-Human + Distractor Meta-Humans (PPQR)}:  
    Two images are selected from the main meta-human, while two are selected from distinct distractor meta-humans.  
    20 sequences are constructed.
    \item \textbf{Main Meta-Human + Distractor Meta-Humans (PQRS)}:  
    One image is selected from the main meta-human, while three are selected from distinct distractor meta-humans.  
    30 sequences are constructed.
\end{enumerate}


\begin{table*}[htbp]
\centering
\begin{tabularx}{\textwidth}{ccYccc}
\toprule
\textbf{Num} & \textbf{Src} & \textbf{Process of Image Sequences Construction} & \textbf{\textit{Cpr}} & \textbf{\textit{cnt}} & \textbf{\textit{Grp}} \\ \hline
\multirow{2}{*}{2} & \multirow{2}{*}{\textbf{PP}} & 2 images from the same person \( P_i \), randomly sampled as \( \mathcal{I}_{P_i} = \{ I_i, I_j \} \), and shuffled. & \multirow{2}{*}{T} & \multirow{2}{*}{2} & \multirow{2}{*}{-} \\ \hline
\multirow{2}{*}{2} & \multirow{2}{*}{\textbf{PQ}} & 1 image \( I_{i} \) from \( \mathcal{I}_{P_i} \) and 1 image \( I_{\neg i} \) from distractor set \( \mathcal{I}_{\neg P_i} \), randomly shuffled. & \multirow{2}{*}{F} & \multirow{2}{*}{1} & \multirow{2}{*}{-} \\ \hline
\multirow{2}{*}{3} & \multirow{2}{*}{\textbf{PPP}} & 3 images from the same person \( P_i \), randomly sampled as \( \mathcal{I}_{P_i} = \{ I_i, I_j, I_k \} \), and shuffled. & \multirow{2}{*}{T} & \multirow{2}{*}{3} & \multirow{2}{*}{-} \\ \hline
\multirow{3}{*}{3} & \multirow{3}{*}{\textbf{PPQ}} & 2 images from the same person \( P_i \), randomly sampled as \( \mathcal{I}_{P_i} = \{ I_i, I_j \} \) and 1 \( I_{\neg i} \) from distractor set \( \mathcal{I}_{\neg P_i} \), randomly shuffled. & \multirow{3}{*}{F} & \multirow{3}{*}{2} & \multirow{3}{*}{[$I_i$, $I_j$]} \\ \hline
\multirow{3}{*}{3} & \multirow{3}{*}{\textbf{PQR}} & 1 image from the same person \( P_i \), randomly sampled as \( \mathcal{I}_{P_i} = \{ I_i \} \) and 2 images \( \{I_{\neg i}, I_{\neg j} \} \) from distractor set \( \mathcal{I}_{\neg P_i} \), randomly shuffled. & \multirow{3}{*}{F} & \multirow{3}{*}{3} & \multirow{3}{*}{[]} \\ \hline
\multirow{2}{*}{4} & \multirow{2}{*}{\textbf{PPPP}} & 4 images from the same person \( P_i \), randomly sampled as \( \mathcal{I}_{P_i} = \{ I_i, I_j, I_k, I_p \} \), and shuffled. & \multirow{2}{*}{T} & \multirow{2}{*}{4} & \multirow{2}{*}{-} \\ \hline
\multirow{3}{*}{4} & \multirow{3}{*}{\textbf{PPPQ}} & 3 images from the same person \( P_i \), randomly sampled as \( \mathcal{I}_{P_i} = \{ I_i, I_j, I_k \} \) and 1 image \( I_{\neg i} \) from distractor set \( \mathcal{I}_{\neg P_i} \), randomly shuffled. & \multirow{3}{*}{F} & \multirow{3}{*}{2} & \multirow{3}{*}{[$I_i$, $I_j$, $I_k$]} \\ \hline
\multirow{3}{*}{4} & \multirow{3}{*}{\textbf{PQQR}} & 2 images from the same person \( P_i \), randomly sampled as \( \mathcal{I}_{P_i} = \{ I_i, I_j \} \) and 2 images \( \{I_{\neg i}, I_{\neg j} \} \) from distractor set \( \mathcal{I}_{\neg P_i} \), randomly shuffled. & \multirow{3}{*}{F} & \multirow{3}{*}{3} & \multirow{3}{*}{[$I_i$, $I_j$]} \\ \hline
\multirow{3}{*}{4} & \multirow{3}{*}{\textbf{PQRV}} & 1 image from the same person \( P_i \), randomly sampled as \( I_i \) and 3 images \( \{I_{\neg i}, I_{\neg j}, I_{\neg k} \} \) from distractor set \( \mathcal{I}_{\neg P_i} \), randomly shuffled. & \multirow{3}{*}{F} & \multirow{3}{*}{3} & \multirow{3}{*}{[]} \\ \bottomrule
\end{tabularx}
\caption{Summary of multi-images sequence construction for Person-centric Cue (PC) tasks.}
\label{tab:sequence construction pc}
\end{table*}

\paragraph{Video Construction.}

\begin{figure}[h]
  \centering
  \includegraphics[width=0.50\textwidth]{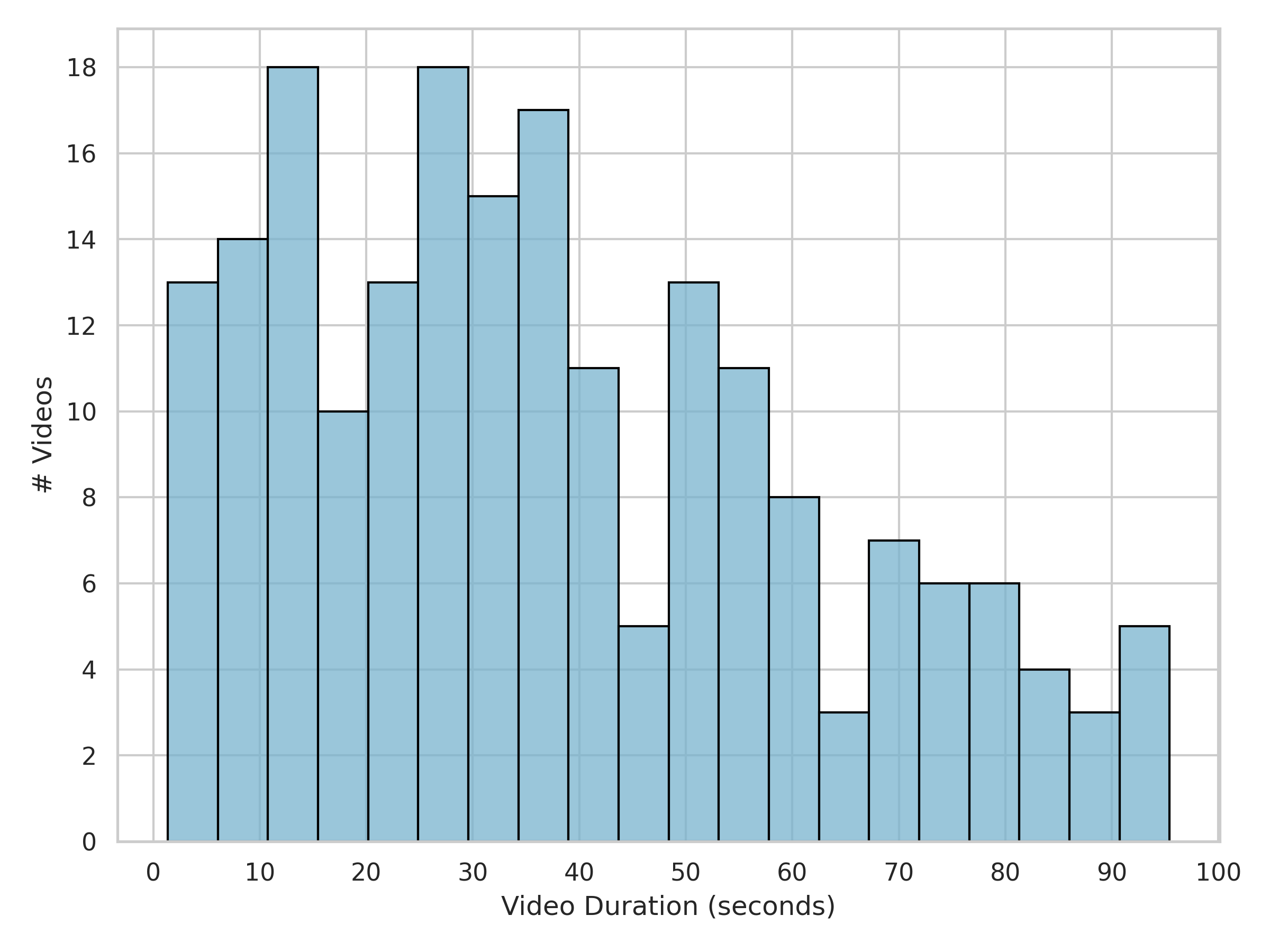} 
  \caption{Distribution of video duration in the subtask of Video Identity Description (VID) in PC.}
  \label{fig:video distribution}
\end{figure}


The video data for this benchmark is manually collected from Shutterstock\footnote{\url{https://www.shutterstock.com}}. We selected ten common activity categories that an individual can perform: \textbf{clean, cook, drink, exercise, listen, play, read, ride, walk, and work}. For each category, we curated \textbf{10 sets of candidate video pairs}, and each set consists of two videos.

To ensure motion consistency and length diversity, we carefully structured the final videos by concatenating clips while keeping the total duration within the \textbf{0-100}s time range. Figure~\ref{fig:video distribution} displays the sketch of concatenated video length distribution. The final compositions followed two formats:
\paragraph{$P$->$\neg P$ format}: A direct concatenation of two distinct clips (same length for each clip).

\paragraph{$P$->$\neg P$->$P$ format}: A sequence where the first clip and the third clip are sampled from the same candidate video, while the second clip is sampled from the second candidate video (same length for the three clips).


Regardless of the different default sampling methods for our baseline models in Table \ref{different_sampling}, both $P$->$\neg P$ and $P$->$\neg P$->$P$ formats ensure that every video clip has frames included while sampling:

\begin{itemize}
    \item \textbf{Uniform Sampling (8/16 frame)}: Each clip contributes a proportionate number of frames based on the total video length. Since in one concatenated video, all the sampled clips are the same length, this method guarantees at least 2 frames for each clip can be sampled as model input frames.
    \item \textbf{FPS Sampling (1fps)}: Since frames are sampled at a fixed rate, the structure of $P$->$\neg P$ and $P$->$\neg P$->$P$ ensures that each clip is present long enough for multiple frames to be captured, regardless of its placement in the sequence.
\end{itemize}

\begin{table}[H]
\centering
{
\begin{tabular}{lcc}
    \toprule
    \textbf{Model Name} & \textbf{Uni} & \textbf{FPS} \\
    \midrule
    LLaVA-OneVision-7B  & \cmark & \xmark \\
    LLaVA-Video-7B      & \cmark & \xmark \\
    LongVA-7B           & \cmark & \xmark \\
    mPLUG-Owl3-7B       & \cmark & \xmark \\
    Qwen2-VL-7B         & \xmark & \cmark \\
    Qwen2.5-VL-7B       & \xmark & \cmark \\
    InternVL2.5-8B      & \cmark & \xmark \\
    InternVL2.5-26B     & \cmark & \xmark \\
    \midrule
    GPT-4o              & \cmark & \xmark \\
    Claude-3.7-sonnet   & \cmark & \xmark \\
    \bottomrule
\end{tabular}
}
    \caption{Comparison of different video sampling methods of VLMs, including Uniform Sampling (Uni) and FPS Sampling (1fps).}
    \label{different_sampling}
\end{table}

Thus, by maintaining the integrity of each clip's temporal structure, both $P$->$\neg P$ and $P$->$\neg P$->$P$ formats effectively ensure that every clip contributes frames to the final sampled frame input for all models.

\paragraph{Question Templates.}

Table~\ref{pc_cpr_tf_example}, Table~\ref{pc_cnt_example}, Table~\ref{pc_grp_example}, and Table~\ref{pc_vid_example} present the detailed standard question templates for the Person-centric Cue task, covering PC-Cpr, PC-Cnt, PC-Grp, and PC-VID.

\begin{table*}[h]
    \begin{tcolorbox}[colframe=black, colback=gray!10!white, coltitle=black, boxrule=0.5mm]
    \textbf{PC-Cpr Positive Question:}  
    
    \textit{
    Judge the following statement based on the images: `The individuals in these images are the same person.' Provide only one correct answer: `T' (True) or `F' (False). Respond with either `T' or `F'.}  
    \\ GT Answer: \textbf{\textcolor{customgreen}{T}}

    \vspace{3mm}
    
    \textbf{PC-Cpr Negative Question:}  
    
    \textit{Judge the following statement based on the images: `The individuals in these images are \textcolor{customred}{not} the same person.' Provide only one correct answer: `T' (True) or `F' (False). Respond with either `T' or `F'.}  
    \\ GT Answer: \textbf{\textcolor{customred}{F}}
    \end{tcolorbox}
    \caption{Question templates used for consistency-pair evaluation in the Comparison (Cpr) subtask of Person-centric Cue (PC).}
    \label{pc_cpr_tf_example}
\end{table*}

\begin{table*}[h]
    \begin{tcolorbox}[colframe=black, colback=gray!10!white, coltitle=black, boxrule=0.5mm]
    \textbf{PC-Cnt Question:}  
    
    \textit{"Answer the following question according to this rule: You only need to provide *ONE* correct numerical answer. For example, if you think the answer is '1', your response should only be '1'. The Question is: How many distinct individuals are in the input images?"}  
    \\ GT Answer: \textbf{2} \textcolor{blue}{(Example Answer)}
    \end{tcolorbox}
    \caption{The question template used for the counting (Cnt) subtask of Person-centric Cue (PC).}
    \label{pc_cnt_example}
\end{table*}

\begin{table*}[h]
    \begin{tcolorbox}[colframe=black, colback=gray!10!white, coltitle=black, boxrule=0.5mm]
    \textbf{PC-Grp Question:}  
    
    \textit{Answer the following question according to this rule: You only need to provide *ONE* correct answer, selecting from the options listed below. For example, if you think the correct answer is `B) 2 and 3', your response should only be `B) 2 and 3'. The Question is: Which images correspond to the same person in the input images? Choices: A) None; B) 2 and 3; C) 1 and 3; D) 1 and 2."}  
    \\ GT Answer: \textbf{D) 1 and 2} \textcolor{blue}{(Example Answer)}
    \end{tcolorbox}
    \caption{The question template used for the grouping (Grp) subtask of Person-centric Cue (PC).}
    \label{pc_grp_example}
\end{table*}

\begin{table*}[h]
    \begin{tcolorbox}[colframe=black, colback=gray!10!white, coltitle=black, boxrule=0.5mm]
    \textbf{PC-VID Question:}  
    
    \textit{"Give a comprehensive description of the whole video, prioritizing details about the individuals in the video."}  
    \end{tcolorbox}
    \caption{The question template used for the Video Identity Description (VID) subtask of Person-centric Cue (PC).}
    \label{pc_vid_example}
\end{table*}

\section{More details on Prompting Approaches}
\label{appendix: prompting approaches}
\subsection{Prompts for LLM-as-Evaluator}

When models answer our free-form PC-VID questions, their responses are evaluated by an evaluator model \citep{fuzzllm} (here GPT-4o) using the scoring prompts detailed in Tables~\ref{score prompt ab} and \ref{score prompt aba}. Specifically, for videos following a \( \mathcal{P} \rightarrow \neg \mathcal{P} \) sequence, GPT-4o assesses whether the model explicitly distinguishes that the first individual (\( \mathcal{P} \)) and the second individual (\( \neg \mathcal{P} \)) are different. In this case, if the model successfully makes this distinction, it receives a score of 1; otherwise, it is given a score of 0.

For videos that exhibit a \( \mathcal{P} \rightarrow \neg \mathcal{P} \rightarrow \mathcal{P} \) (PQP) pattern, the evaluation is more nuanced. The evaluator model (GPT-4o) checks two aspects: (1) whether the model correctly identifies that there are two distinct individuals (i.e., \( \mathcal{P} \) and \( \neg \mathcal{P} \)), and (2) whether the model explicitly recognizes that the final appearance belongs to the same individual as the first (\( \mathcal{P} \)). A perfect identification of both aspects yields a score of 2, while correctly distinguishing the individuals without explicitly linking the final appearance to the first results in a score of 1. If the model fails to distinguish between the individuals, a score of 0 is assigned.




\begin{table*}[htbp]
    \begin{tcolorbox}[colframe=black, colback=gray!10!white, coltitle=black, boxrule=0.5mm]
    \textbf{\#Task}
    
    You are evaluating a model's ability to accurately distinguish between two different individuals, P and Q, who appear sequentially in a video (first P, then Q). Given a description, your task is to determine if the model explicitly identifies that the first person (P) and the second person (Q) are different individuals.
    
    \textbf{\#Return Format}
    
    You only need return a number after "Score:". If you think the model correctly identifies that the two appearances belong to different individuals, return "Score: 1". If you think the model fails to explicitly state that there are two different individuals, return "Score: 0".
    
    \textbf{\#Description}
    
    \textcolor{teal}{\textbf{<Model's Description>}}
    \end{tcolorbox}
    \caption{Scoring prompt for \textit{VID} (when video belongs to category of $P$->$\neg P$).}
    \label{score prompt ab}
\end{table*}

\begin{table*}[htbp]
    \begin{tcolorbox}[colframe=black, colback=gray!10!white, coltitle=black, boxrule=0.5mm]
    \textbf{\#Task}
    
    You are evaluating a model's ability to accurately distinguish between two different individuals, P and Q, who appear sequentially in a video following an PQP pattern (first P, then Q, then P again). Given a description, your task is to determine whether the model explicitly identifies that: (1) P and Q are different individuals, and (2) The person in the final scene is the same as the first (P).
    
    \textbf{\#Return Format}
    
    You only need return a number after "Score:". 
    
    (1) If the model correctly describes that the video follows an PQP sequence, explicitly recognizing that the first and last appearances belong to the same person (P), while the middle appearance is a different person (Q), return "Score: 2".
    
    (2) If the model correctly identifies that there are two different people in the video (P and Q) but does not explicitly mention that the last scene returns to P, return "Score: 1".

    (3) If the model fails to recognize that two different individuals appear (e.g., treats all appearances as the same person or does not distinguish between P and Q), return "Score: 0".
    
    \textbf{\#Description}
    
    \textcolor{teal}{\textbf{<Model's Description>}}
    \end{tcolorbox}
    \caption{Scoring prompt for \textit{VID} (when video belongs to category of $P$->$\neg P$->$P$).}
    \label{score prompt aba}
\end{table*}

\subsection{Prompting Approaches for Probing on VLM2-Bench}
\paragraph{CoT (CoT-normal).}
The normal version of the Chain-of-Thought prompt is shown in Table~\ref{cot prompt}. We simply require the model to think 'step-by-step' to ensure self-reflection and self-correction, as well as the transparent thinking process.

\paragraph{CoT-special for GC.}
Table~\ref{cot-special} shows a special version of the Chain-of-Thought prompt. According to the task features, we carefully analyze how a human being approaches and visually links matching cues for questions in GC, then curate this prompt as an imitation of the human visual linking process.

\begin{figure*}[htbp]
\centering
\includegraphics[width=0.95\textwidth]{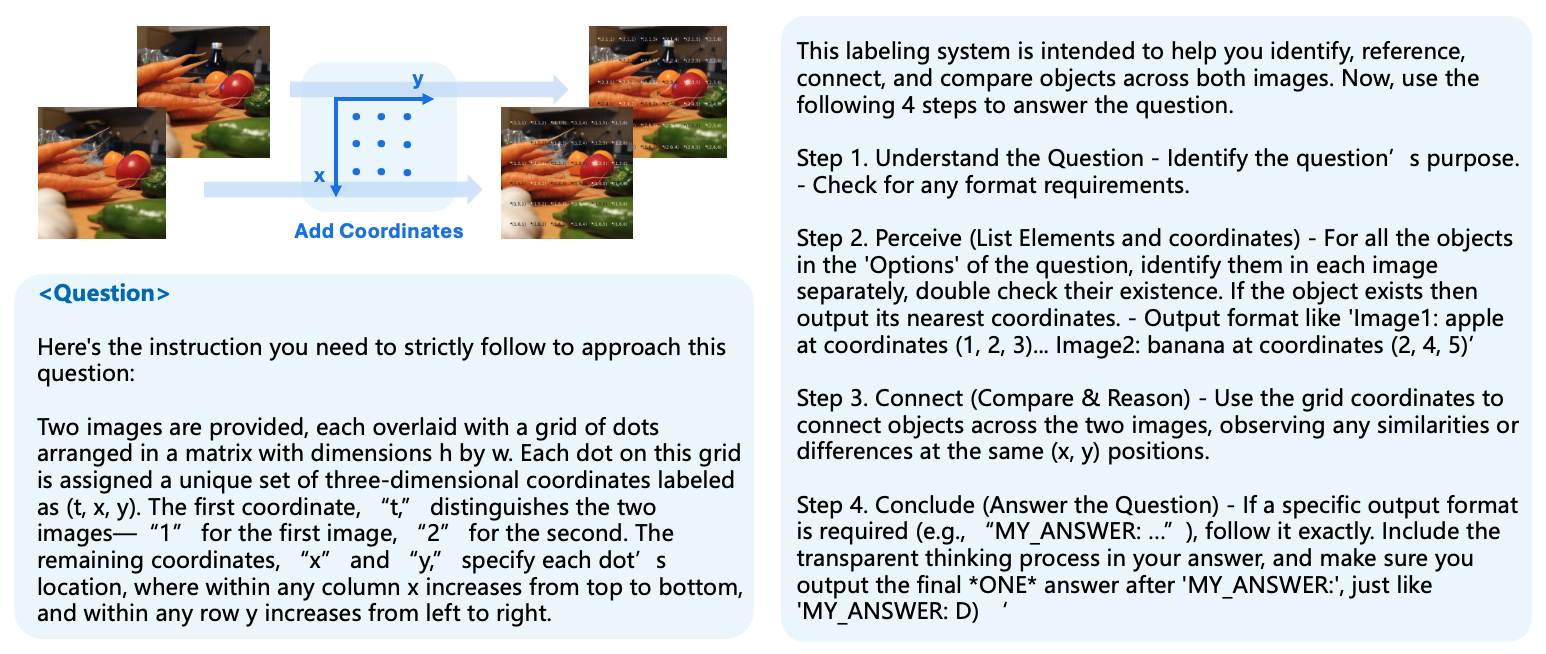} 
\caption{An illustration of how VP-grid works for GC.}
\label{fig:vp-grid}
\end{figure*}

\paragraph{VP-grid for GC.}
Figure~\ref{fig:vp-grid} displays a complete version of Visual Prompting with Grid assistance (VP-grid). Here we follow ~\citep{lei2024scaffoldingcoordinatespromotevisionlanguage} to print a set of dot matrix onto the input image, accompanied by the image order dimension concatenated with Cartesian coordinates as (\textit{image order index}, \textit{colum index)}, \textit{row index}). In the detailed textual prompt design, we also integrated references and explanations for the grids, allowing VLMs to leverage this visual assistance as spatial and visual matching references.

\begin{figure*}[h]
\centering
\includegraphics[width=0.95\textwidth]{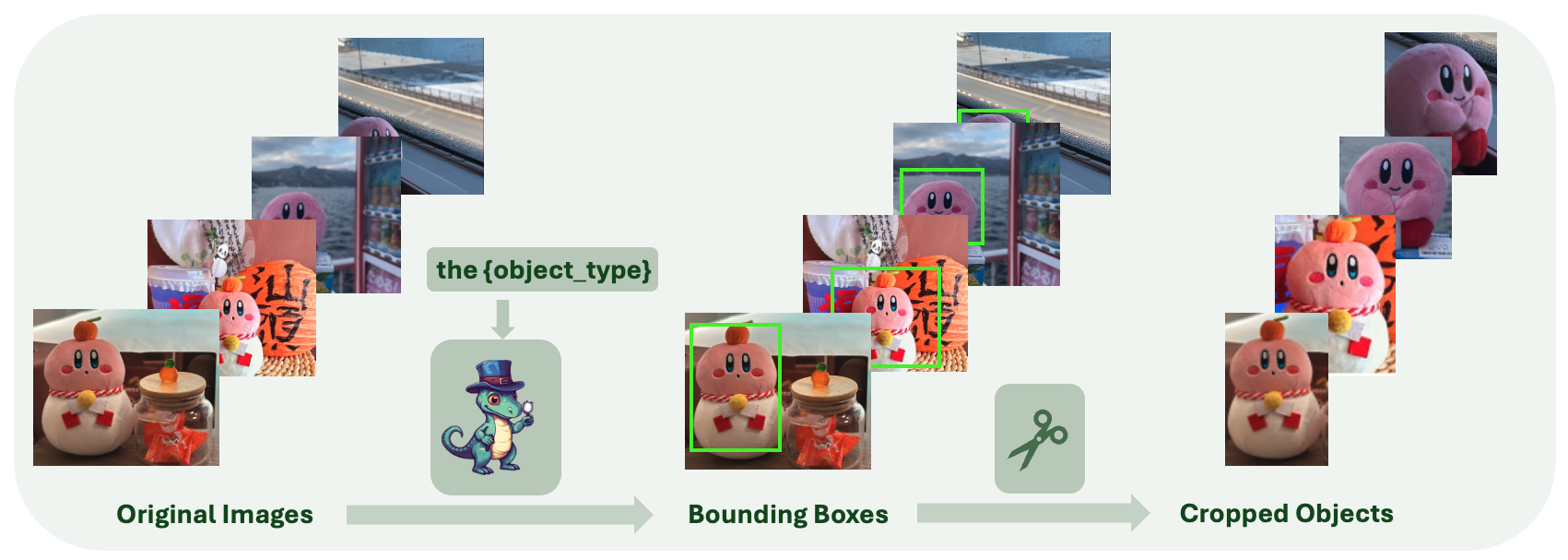} 
\caption{An illustration of how VP-zoom-o works for OC.}
\label{fig:VP-zoom-o}
\end{figure*}

\paragraph{VP-zoom-o for OC.}
In Figure~\ref{fig:VP-zoom-o}, we demonstrate the visual prompting process for OC. We leverage the Grounded-SAM \citep{grounded-sam} model to detect bounding boxes for objects based on their types then crop the ``zoomed-in'' objects as the image input for further VQA pairs.

\begin{figure*}[h]
\centering
\includegraphics[width=0.95\textwidth]{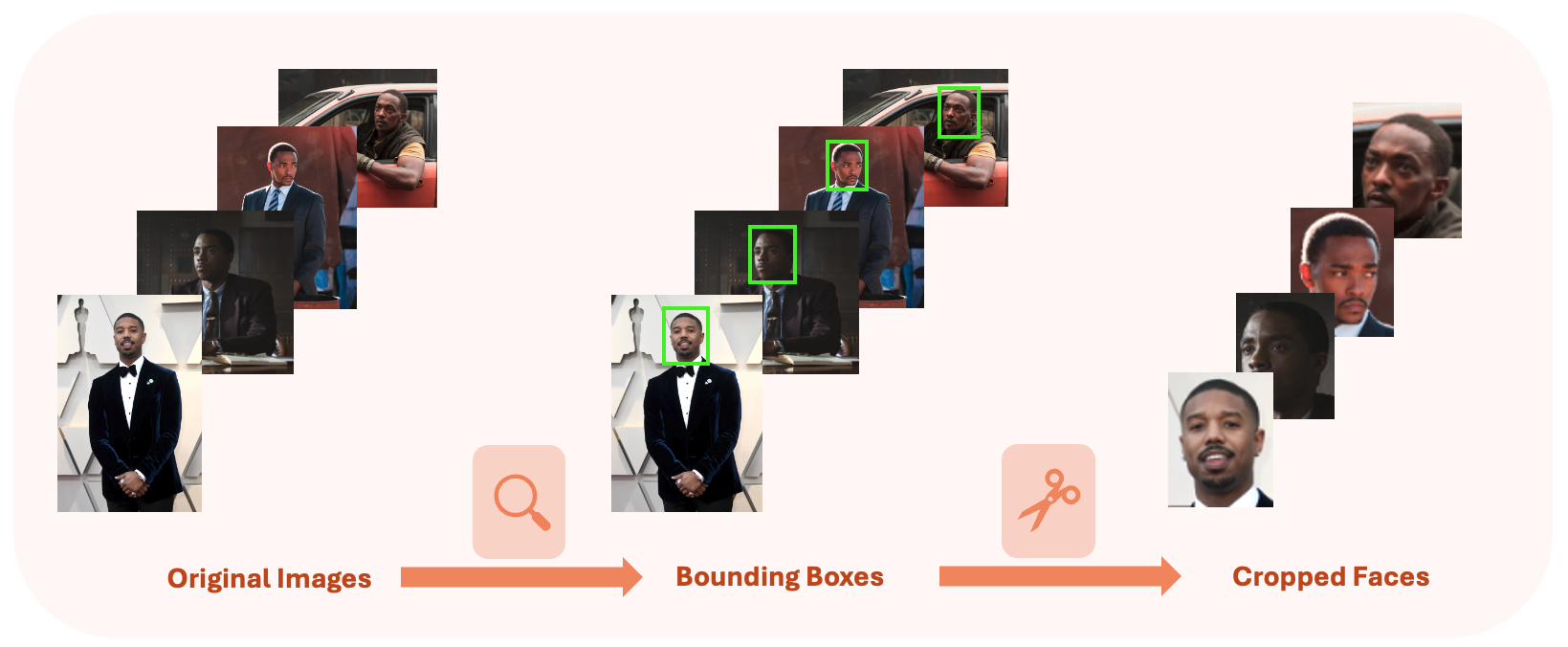} 
\caption{An illustration of how VP-zoom-p works for PC.}
\label{fig:vp-zoom-p}
\end{figure*}

\paragraph{VP-zoom-p for PC.}
The visual prompting process in similar to that of OC (Figure~\ref{fig:vp-zoom-p}). We use a face detection model~\citep{facedetector} to ``zoom in'' on the individual's face and occlude other irrelevant information.

\begin{table*}[htbp]
    \begin{tcolorbox}[colframe=black, colback=gray!10!white, coltitle=black, boxrule=0.5mm]
    \textcolor{blue!70}{\textbf{<Question>}}
    
    Let's think `step by step' to answer this question, you need to output the thinking process of how you get the answer.
    \end{tcolorbox}
    \caption{CoT prompt for GC (here we denote as CoT-normal to distinguish it from the CoT-special in Table~\ref{cot-special} that specifically designed for GC), OC, and PC.}
    \label{cot prompt}
\end{table*}

\begin{table*}[htbp]
    \begin{tcolorbox}[colframe=black, colback=gray!10!white, coltitle=black, boxrule=0.5mm]
    \textcolor{blue!70}{\textbf{<Question>}}
    
    Use the following 4 steps to answer the question:
    \vspace{0.5em}
    
    \textbf{Step 1. Understand the Question} \\
    - Identify the question's purpose.\\
    - Check for any format requirements.\\
    
    \textbf{Step 2. Perceive (List Elements)} \\
    - List every details in each image respectively. \\
    - Note positions and attributes of elements.\\
    
    \textbf{Step 3. Connect (Compare \& Reason)} \\
    - Compare corresponding elements in each image.\\
    - List all the unchanged elements and the changed element.\\
    
    \textbf{Step 4. Conclude (Answer the Question)} \\
    \end{tcolorbox}
    \caption{CoT-special specifically designed for GC.}
    \label{cot-special}
\end{table*}

\section{Case Study}
\label{Appendix: case study}

This section focuses on how various prompting techniques influence model performance, highlighting their successes and limitations across different models.

\subsection{Case for CoT-special prompting in General Cue (GC) Task}

We observe that the CoT-special prompt boosts InternVL2.5-8B's performance by over 25\% than the standard query in both Matching and Tracking tasks for General Cue. While for the traditional CoT-normal prompting technique, this boost is only 13\%. The CoT-special prompt (Table~\ref{cot-special}) directs the model through four explicit steps: understanding the question, perceiving (listing elements), connecting (comparing and reasoning), and concluding. This structured approach mirrors the human process of visual matching and is effective even for a rather smaller model like InternVL2.5-8B, which might otherwise struggle with the ambiguity of a complex generic step-by-step instruction (which we will discuss later in the next Subsection~\ref{vp-decrease case study}).

\begin{figure*}[htbp]
  \centering
  \includegraphics[width=0.75\textwidth]{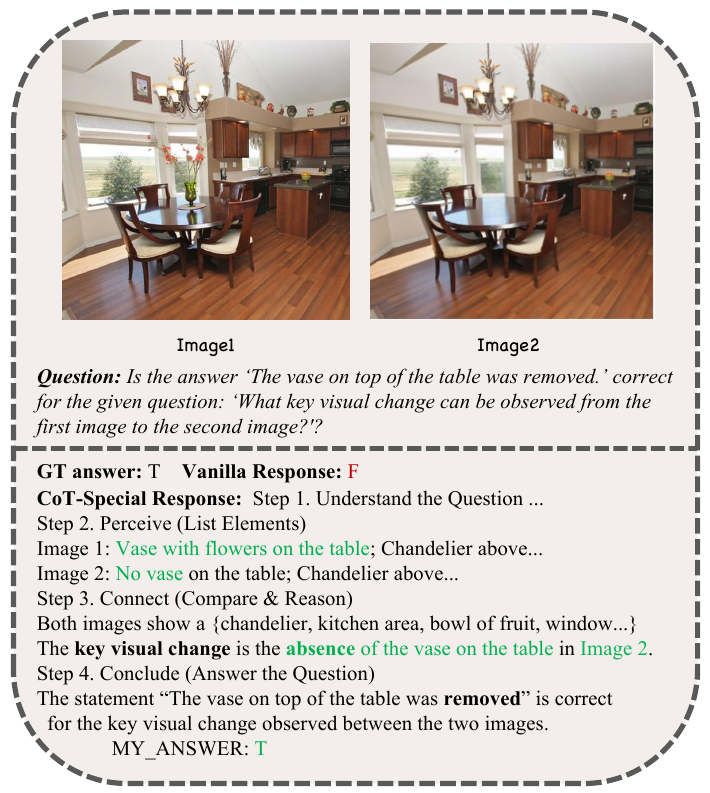} 
  \caption{Case study illustrating how CoT-special improves performance of the subtask of Tracking (Trk) in General Cue (GC). The model, InternVL2.5-8B, demonstrates a step-by-step reasoning process: In Step 2, it identifies key details such as ``Vase with flowers on the table" and "Chandelier above" in Image 1, while noting the absence of the vase in Image 2. In Step 3, it compares the images, recognizing that while many elements remain unchanged (e.g., chandelier, kitchen area, fruit bowl, window), the vase's removal is the primary difference. In Step 4, the model concludes that the statement "The vase on top of the table was removed" accurately reflects the visual change, leading to the correct answer.}

  \label{fig:cot-special increase}
\end{figure*}

For example, in the provided InternVL2.5-8B response Figure~\ref{fig:cot-special increase}, the model correctly executes the following: In Step 2, it identifies critical details such as "Vase with flowers on the table" and "Chandelier above" in Image 1, while noting the absence of the vase in Image 2. In Step 3, it systematically compares the two images, highlighting that while many elements remain unchanged (e.g., the chandelier, kitchen area, bowl of fruit, window), the removal of the vase is the key difference. Finally, in Step 4, the model concludes that the statement "The vase on top of the table was removed" accurately describes the visual change, thereby arriving at the correct answer.

This detailed, multi-step breakdown not only ensures that all pertinent visual cues are captured and processed but also reduces errors by structuring the logical flow of reasoning. The CoT-special prompt's explicit instructions help InternVL2.5-8B align visual information with textual descriptions more effectively, thus enhancing overall performance. Compared to the less specific CoT-normal prompt—which may leave the model with gaps in reasoning—the CoT-special prompt provides clear, task-specific guidance that is essential for complex visual reasoning tasks, as evidenced by the substantial performance improvement.

\subsection{Case for VP-grid in General Cue Task}
\label{vp-decrease case study}
The VP-grid (Visual Prompting with Grid assistance) method enhances visual matching in General Cue tasks by overlaying a dot matrix grid onto the input image. Each dot is annotated with a three-dimensional coordinate tuple, \((\textit{image order index}, \textit{column index}, \textit{row index})\), where the first dimension distinguishes the sequence of images (e.g., the first image is indexed as 1 and the second as 2). This grid is further supported by detailed textual descriptions that clarify the coordinate system, enabling Vision-Language Models (VLMs) to use these cues for spatial and visual matching.

\begin{figure*}[h]
  \centering
  \includegraphics[width=0.75\textwidth]{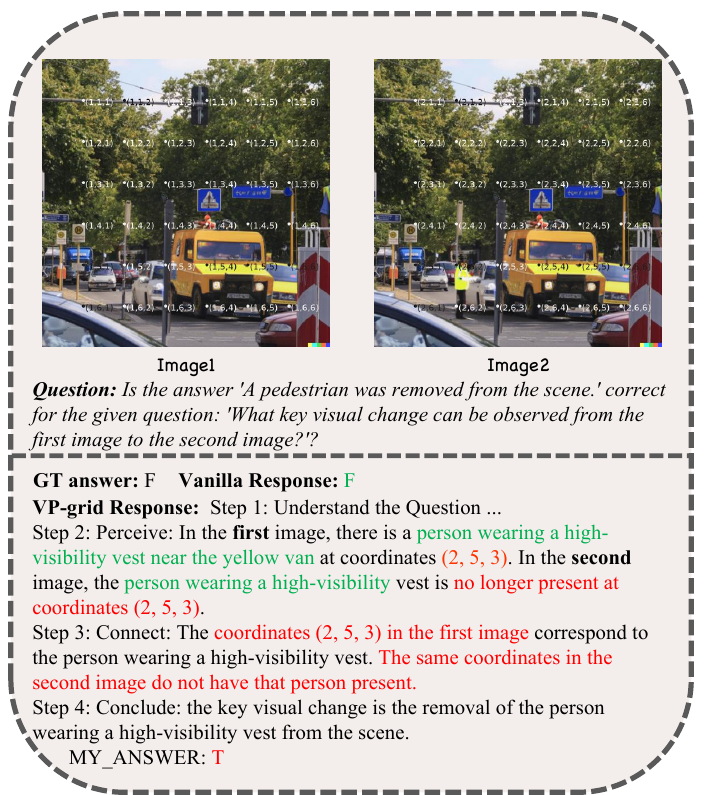} 
  \caption{Case study illustrating why VP-grid leads to performance degradation in Qwen2.5-VL-7B. The model correctly identifies visual elements (e.g., a pedestrian with a high-visibility vest at coordinates \((2,5,3)\)) but fails to interpret the image sequence correctly. It mistakenly associates the coordinates with the first image instead of the second, despite the explicit definition in the textual prompt. This misinterpretation results in incorrect visual cue linking and faulty reasoning, highlighting the model's difficulty in handling structured spatial instructions and visual prompts.}
  \label{fig:vp-grid decrease}
\end{figure*}

\paragraph{An example failure case in VP-grid.}
However, this approach does not yield consistent improvements across all models. For instance, the Qwen2.5-VL-7B model demonstrates a significant performance drop—nearly 20\%—when using VP-grid. An example failure case is in Figure ~\ref{fig:vp-grid decrease}. Our analysis reveals that although the model correctly identifies visual elements (e.g., a pedestrian with a high-visibility vest at coordinates \((2,5,3)\)), it fails to properly interpret the image sequence. Specifically, the model incorrectly associates the coordinates \((2,5,3)\) with the first image, rather than the second, despite the explicit definition provided in the textual prompt. This misinterpretation leads to erroneous linking of visual matching cues and subsequent faulty reasoning. We suspect that the underlying issue is the limited semantic comprehension capability of the relatively smaller 7B model, which struggles with complex, predefined spatial instructions and visual assistance.

\begin{figure*}[h]
  \centering
  \includegraphics[width=0.75\textwidth]{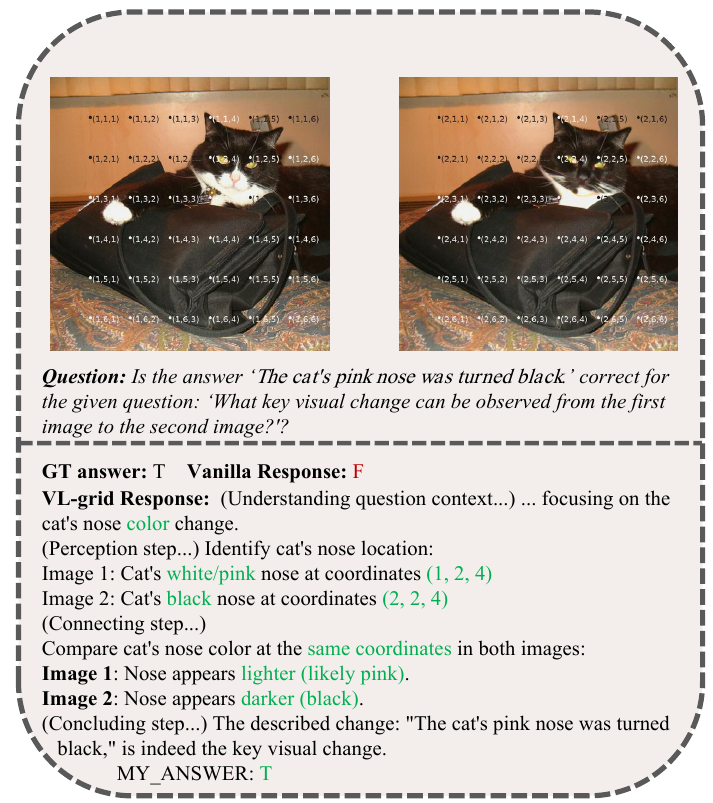} 
  \caption{Case study demonstrating why VP-grid leads to performance improvement for GPT-4o. Unlike models that often misinterpret or overlook spatial cues, GPT-4o effectively uses VP-grid to align visual and textual information. In the example shown in Figure~\ref{fig:vp-grid increase}, the model correctly identifies the cat's nose at coordinates \((1,2,4)\) in the first image and \((2,2,4)\) in the second, accurately capturing the visual change in the attribute (from a lighter pink to a darker black). This success highlights GPT-4o’s ability to handle structured spatial prompts and improve performance through visual prompting.}
  \label{fig:vp-grid increase}
\end{figure*}

\paragraph{An example of success case in VP-grid.} In contrast to models that often misinterpret or neglect spatial cues provided by VP-grid—leading to errors such as mismatching image indices—GPT-4o successfully leverages these visual prompts to achieve correct visual-textual alignment. In the example at Figure~\ref{fig:vp-grid increase}, the model identifies the cat's nose at coordinates \((1,2,4)\) in the first image and at \((2,2,4)\) in the second image, enabling it to accurately capture the change in the visual attribute (from a lighter pink to a darker black).

This success stems from several key aspects of GPT-4o's processing capabilities:
\begin{enumerate}
    \item \textbf{Precise Disambiguation of Image Order:} The VP-grid explicitly encodes image order, which GPT-4o uses to differentiate between multiple images. This prevents the common error of conflating spatial information from distinct images—a problem seen in smaller models.
    \item \textbf{Robust Visual Matching in space:} With clear coordinate annotations, the model effectively locates and compares the same physical regions across images. In this case, the exact correspondence between the cat's nose in different images is recognized, which is crucial for detecting subtle visual changes.
    \item \textbf{Structured Reasoning Process:} GPT-4o adheres to a well-defined reasoning sequence in our textual guidance(perception, connection, and conclusion). By systematically linking the provided grid coordinates with the textual descriptions, it is able to deduce the key visual change accurately.
\end{enumerate}

\paragraph{Implications on Model Scale.} Our analysis suggests that the enhanced performance of GPT-4o with VP-grid can be attributed to its larger model capacity. Although the detailed architecture of GPT-4o is proprietary, its ability to process complex multi-modal prompts implies that:
\begin{itemize}
    \item \textbf{Enhanced Semantic Understanding:} Larger models are inherently better at comprehending intricate, structured prompts that combine visual and textual information. This results in a more nuanced interpretation of spatial cues.
    \item \textbf{Superior Visual-Textual Alignment:} With greater capacity, GPT-4o can integrate and correlate the detailed spatial data (visual assistance) from the VP-grid with the corresponding textual descriptions, minimizing the risk of mis-association or errors.
    \item \textbf{Effective Handling of Complexity:} The advanced reasoning capabilities of larger models enable them to navigate the additional complexity introduced by VP-grid without suffering from the side effects seen in smaller models. This ensures that the additional spatial guidance improves performance rather than causing confusion.
\end{itemize}

The success of GPT-4o in utilizing the VP-grid approach demonstrates that model scale plays a critical role in effectively integrating complex visual and textual cues. By accurately disambiguating image order and performing precise spatial matching, GPT-4o not only avoids the pitfalls encountered by smaller models but also benefits significantly from the additional visual assistance, leading to an overall performance improvement of approximately 10\%.

\subsection{Case for CoT prompting in Object-centric Cue Task}

The task design for Object-centric cue (OC) and person-centric cue (PC) requires multiple images (more than 2) as sequence input. We observe that, unlike General Cue (GC) tasks where models are required to link instance-level cues, OC tasks demand that models group similar objects based on fine-grained visual features. As illustrated in Figure~\ref{fig:oc-analysis}, models using the CoT approach sometimes struggle to provide a comprehensive overview of vision-based cues across a sequence of images. 

\begin{figure*}[h]
  \centering
  \includegraphics[width=0.75\textwidth]{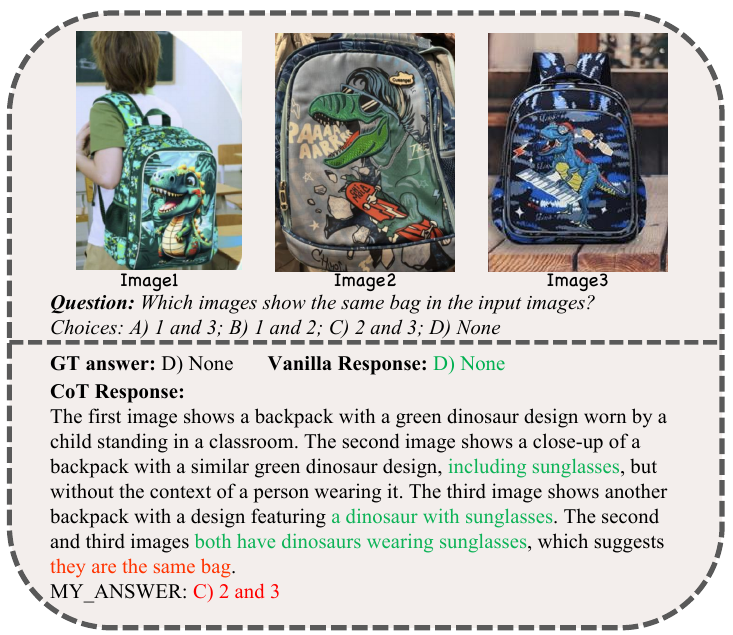} 
  \caption{Case study illustrating why CoT leads to performance degradation. In the example shown in Figure~\ref{fig:oc_cot_normal_decrease}, InternVL2.5-26B's response correctly identifies that no grouping occurs for the same meta-object in the sequence, with the correct answer being `D) None'. However, in the CoT response, the model incorrectly selects option C) 2 and 3. While it correctly states that ``the second and third images both have dinosaurs wearing sunglasses," the lack of detailed analysis leads to an inaccurate conclusion. A closer examination reveals a key difference between the images—the dinosaur in image 3 is holding a keyboard instead of a skateboard, which should have prevented the grouping of the two images. This highlights the importance of providing more detailed and unambiguous cues in CoT reasoning.}

  \label{fig:oc_cot_normal_decrease}
\end{figure*}

A detailed case in Figure~\ref{fig:oc_cot_normal_decrease} is provided by InternVL2.5-26B's response. The ground truth and Vanilla responses correctly identify that there is no grouping for the same meta-object in the sequence, with the answer `D) None'. In the CoT response, the model states: "The second
and third images \textcolor{customgreen}{both have dinosaurs wearing sunglasses}". Although the description here is true, its ambiguity and lack of detailed coverage lead the model to incorrectly select option C) 2 and 3, rather than the correct option D) None. Because if we take a closer look at the design on the backpack in image 3, the dinosaur with sunglasses is actually holding a keyboard instead of a skateboard in image 2. This is a distractive visual matching cue we intend to capture during the distractor meta-object selection. This major difference should have prevented models from grouping image 2 and image 3 together.

According to our findings, this misgrouping occurs for two main reasons:
\begin{enumerate}
    \item \textbf{Insufficient Overview of Visual Cues:} The CoT prompt does not force the model to systematically verify all critical details across multiple images. As a result, the model overlooks nuanced differences, such as the design discrepancy on the backpack in image 3, where the dinosaur holds a keyboard rather than a skateboard.
    \item \textbf{Variability in Descriptive Language:} The open-ended language generated by the CoT approach can lead to inconsistent descriptions. In this case, the model generalized the visual cue of a "dinosaur design" without capturing the specific attribute (i.e., the object the dinosaur is holding), which is crucial for correct grouping.
\end{enumerate}

Thus, the lack of structured guidance in the CoT prompt leads to the dropping or misinterpretation of critical cues, resulting in incorrect grouping decisions for multi-image sequences in OC tasks. This analysis underscores the importance of more detailed structured intermediate reasoning strategies, such as those provided by a tailored CoT-special prompt, to ensure that all relevant visual details are captured and compared accurately.

\end{document}